\pdfoutput=1
\documentclass[runningheads]{llncs}
\usepackage{graphicx}
\usepackage{amsmath,amssymb} % define this before the line numbering.
\usepackage{color}
\usepackage{booktabs}

\usepackage[binary-units]{siunitx}

\usepackage{float}

\begin{document}
\title{Fast Perceptual Image Enhancement} 
% Replace with your title

\titlerunning{Fast Perceptual Image Enhancement}
% Replace with a meaningful short version of your title
%
\author{Etienne de Stoutz\orcidID{0000-0001-5439-3290} \and
Andrey Ignatov\orcidID{0000-0003-4205-8748} \and
Nikolay Kobyshev\orcidID{0000-0001-6456-4946} \and
Radu Timofte \orcidID{0000-0002-1478-0402}\and
Luc Van Gool\orcidID{0000-0002-3445-5711}}
%
%Please write out author names in full in the paper, i.e. full given and family names. 
%If any authors have names that can be parsed into FirstName LastName in multiple ways, please include the correct parsing, in a comment to the volume editors:
%\index{de Stoutz, Etienne}
%\index{Van Gool, Luc}
%(Do not uncomment it, because you may introduce extra index items if you do that, we will use scripts for introducing index entries...)
\authorrunning{de Stoutz, E., Ignatov, A., Kobyshev, N., Timofte, R., Van Gool, L.}
% Replace with shorter version of the author list. If there are more authors than fits a line, please use A. Author et al.
%

\institute{ETH Zurich}
\maketitle              % typeset the header of the contribution
\begin{abstract}
The vast majority of photos taken today are by mobile phones. While their quality is rapidly growing, due to physical limitations and cost constraints, mobile phone cameras struggle to compare in quality with DSLR cameras. This motivates us to computationally enhance these images. We extend upon the results of Ignatov~\textit{et al.}, where they are able to translate images from compact mobile cameras into images with comparable quality to high-resolution photos taken by DSLR cameras. However, the neural models employed require large amounts of computational resources and are not lightweight enough to run on mobile devices. We build upon the prior work and explore different network architectures targeting an increase in image quality and speed. With an efficient network architecture which does most of its processing in a lower spatial resolution, we achieve a significantly higher mean opinion score (MOS) than the baseline while speeding up the computation by 6.3$\times$ on a consumer-grade CPU. This suggests a promising direction for neural-network-based photo enhancement using the phone hardware of the future.
\end{abstract}

\section{Introduction}

The compact camera sensors found in low-end devices such as mobile phones have come a long way in the past few years. Given adequate lighting conditions, they are able to reproduce unprecedented levels of detail and color. Despite their ubiquity, being used for the vast majority of all photographs taken worldwide, they struggle to come close in image quality to DSLR cameras. These professional grade instruments have many advantages including better color reproduction, less noise due to larger sensor sizes, and better automatic tuning of shooting parameters.

Furthermore, many photographs were taken in the past decade using significantly inferior hardware, for example with early digital cameras or early 2010s smartphones. These do not hold up well to our contemporary tastes and are limited in artistic quality by their technical shortcomings.

The previous work by Ignatov~\textit{et al.}~\cite{ignatov2017dslr} that this paper is based upon proposes a neural-network powered solution to the aforementioned problems. They use a dataset comprised of image patches from various outdoor scenes simultaneously taken by cell phone cameras and a DSLR. They pose an image translation problem, where they feed the low-quality phone image into a residual convolutional neural net (CNN) model that generates a target image, which, when the network is trained, is hopefully perceptually close to the high-quality DSLR target image.

In this work, we take a closer look at the problem of translating poor quality photographs from an iPhone 3GS phone into high-quality DSLR photos, since this is the most dramatic increase in quality attempted by Ignatov~\textit{et al.}~\cite{ignatov2017dslr}. The computational requirements of this baseline model, however, are quite high ($\SI{20}\second$ on a high-end CPU and $\SI{3.7}{\giga\byte}$ of RAM for a HD-resolution image). Using a modified generator architecture, we propose a way to decrease this cost while maintaining or improving the resulting image quality.

\section{Related Work}

A considerable body of work is dedicated to automatic photo enhancement. However, it traditionally only focused on a specific subproblem, such as super-resolution, denoising, deblurring, or colorization. All of these subproblems are tackled simultaneously when we generate plausible high-quality photos from low-end ones. Furthermore, these older works commonly train with artifacts that have been artificially applied to the target image dataset. Recreating and simulating all the flaws in one camera given a picture from another is close to impossible, therefore in order to achieve real-world photo enhancement we use the photos simultaneously captured by a capture rig from Ignatov~\textit{et al.}~\cite{ignatov2017dslr}. Despite their limitations, the related works contain many useful ideas, which we briefly review in this section.

\textit{Image super-resolution} is the task of increasing the resolution of an image, which is usually trained with down-scaled versions of the target image as inputs. Many prior works have been dedicated to doing this using CNNs of progressively larger and more complex nature \cite{dong2014learning,kim2016accurate,mao2016image,shi2016real,timofte2017ntire,Timofte_2018_CVPR_Workshops}. Initially, a simple pixel-wise mean squared error (MSE) loss was often used to guarantee high fidelity of the reconstructed images, but this often led to blurry results due to uncertainty in pixel intensity space. Recent works~\cite{blau2018pirm} aim at perceptual quality and employ losses based on VGG layers~\cite{johnson2016perceptual}, and generative adversarial networks (GANs)~\cite{goodfellow2014generative,ledig2017photo}, which seem to be well suited to generating plausible-looking, realistic high-frequency details.

In \textit{image colorization}, the aim is to hallucinate color for each pixel, given only its luminosity. It is trained on images with their color artificially removed. Isola~\textit{et al.}~\cite{isola2017image} achieve state of the art performance using a GAN to solve the more general problem of image-to-image translation.

\textit{Image deblurring and dehazing} aim to remove optical distortions from photos that have been taken out of focus, while the camera was moving, or of faraway geographical or astronomical features. The neural models employed are CNNs, typically trained on images with artificially added blur or haze, using a MSE loss function~\cite{ren2016single,li2017aod,ling2016learning,hradivs2015convolutional,cai2016dehazenet}. Recently, datasets with both hazy and haze-free images were introduced~\cite{Ancuti_2018_CVPR_Workshops} and solutions such as the one of Ki~\textit{et al.}~\cite{ki2018fully} were proposed, which use a GAN, in addition to L1 and perceptual losses. Similar techniques are effective for \textit{image denoising} as well~\cite{zhang2016fast,zhang2017beyond,yang2016joint,svoboda2016compression}.

\subsection{General Purpose Image-to-Image Translation and Enhancement}
The use of GANs has progressed towards the development of general purpose image-to-image translation. Isola~\textit{et al.}~\cite{isola2017image} propose a conditional GAN architecture for paired data, where the discriminator is conditioned on the input image. Zhu~\textit{et al.}~\cite{zhu2017unpaired} relax this requirement, introducing the cycle consistency loss which allows the GAN to train on unpaired data. These two approaches work on many surprising datasets, however, the image quality is too low for our purpose of photo-realistic image enhancement. This is why Ignatov \textit{et al.} introduce paired ~\cite{ignatov2017dslr} and unpaired~\cite{ignatov2017wespe} GAN architectures that are specially designed for this purpose.

\subsection{Dataset}
The DPED dataset~\cite{ignatov2017dslr} consists of photos taken simultaneously by three different cell phone cameras, as well as a Canon 70D DSLR camera. In addition, these photographs are aligned and cut into 100x100 pixel patches, and compared such that patches that differ too much are rejected. In this work, only the iPhone 3GS data is considered. This results in 160k pairs of images.
\subsection{Baseline}
\begin{figure}[htb]
\includegraphics[width=\linewidth]{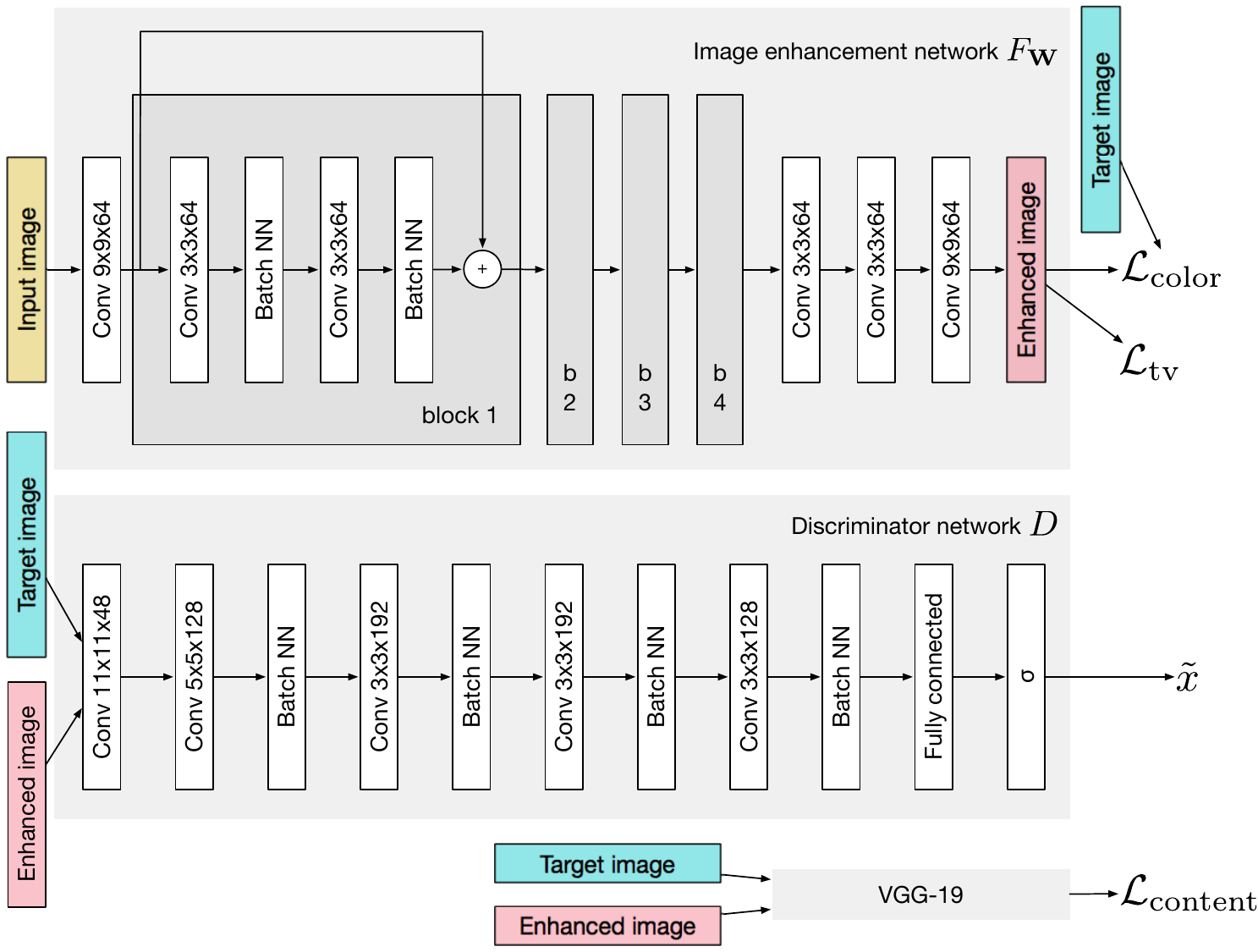}
\caption{The overall architecture of the DPED baseline~\cite{ignatov2017dslr}}
\end{figure}
As a baseline, the residual network with 4 blocks and 64 channels from Ignatov \textit{et al.}~\cite{ignatov2017dslr} is used.

Since using a simple pixel-wise distance metric does not yield the intended perceptual quality results, the output of the network is evaluated using four carefully designed loss functions. 

The generated image is compared to the target high-quality DSLR image using the color loss and the content loss. The same four losses and training setup as the baseline are also used by us in this work.

\subsubsection{Color Loss.}

The color loss is computed by applying a Gaussian blur to both source and target images, followed by a MSE function. Let $X$ and $Y$ be the original images, then $X _ { b }$ and $Y _ { b }$ are their blurred versions, using
\begin{equation}
X _ { b } ( i , j ) = \sum _ { k , l } X ( i + k , j + l ) \cdot G ( k , l ),
\end{equation}
where $G$ is the 2D Gaussian blur operator
\begin{equation}
G ( k , l ) = A \exp \left( - \frac { \left( k - \mu _ { x } \right) ^ { 2 } } { 2 \sigma _ { x } } - \frac { \left( l - \mu _ { y } \right) ^ { 2 } } { 2 \sigma _ { y } } \right).
\end{equation}
The color loss can then be written as
\begin{equation}
\mathcal { L } _ { \mathrm { color } } ( X , Y ) = \left\| X _ { b } - Y _ { b } \right\| _ { 2 } ^ { 2 }.
\end{equation}
We use the same parameters as defined in \cite{ignatov2017dslr}, namely $A = 0.053 , \mu _ { x , y } = 0$, and $\sigma _ { x , y } = 3$.

\subsubsection{Content Loss.}
The content loss is computed by comparing the two images after they have been processed by a certain number of layers of VGG-19. This is superior to a pixel-wise loss such as per-pixel MSE, because it closely resembles human perception~\cite{ignatov2017dslr,zhang2018unreasonable}, abstracting away such negligible details as a small shift in pixels, for example. It is also important because it helps preserve the semantics of the image. It is defined as
\begin{equation}
\mathcal { L } _ { \mathrm { content } } = \frac { 1 } { C _ { j } H _ { j } W _ { j } } \left\| \psi _ { j } \left( F _ { \mathrm { w } } \left( I _ { s } \right) \right) - \psi _ { j } \left( I _ { t } \right) \right\|
\end{equation}
where $\psi _ { j } (\cdot)$ is the feature map of the VGG-19 network after its $j$-th convolutional layer, $C _ { j } , H _ { j } \text {, and } W _ { j }$ are the number, height, and width of this map, and $F _ { \mathbf { W } } \left( I _ { s } \right)$ denotes the enhanced image.

\subsubsection{Texture Loss.}
One important loss which technically makes this network a GAN is the texture loss~\cite{ignatov2017dslr}. Here, the output images are not directly compared to the targets, instead, a discriminator network is tasked with telling apart real DSLR images from fake, generated ones. During training, its weights are optimized for maximum discriminator accuracy, while the generator's weights are optimized in the opposite direction, to try to minimize the discriminator's accuracy, therefore producing convincing fake images.

Before feeding the image in, it is first converted to grayscale, as this loss is specifically targeted on texture processing. It can be written as
\begin{equation}
\mathcal{L}_{\text{texture}} = -\sum_{i} \log{D(F_{\textbf{W}}(I_s), I_t)},
\end{equation}
where $F_{\textbf{W}}$ and $D$ denote the generator and discriminator networks, respectively.

\subsubsection{Total Variation Loss.}
A total variation loss is also included, so as to encourage the output image to be spatially smooth, and to reduce noise.
\begin{equation}
\mathcal{L}_{\text{tv}} = \frac{1}{C H W} \|\nabla_x F_{\textbf{W}}(I_s) + \nabla_y F_{\textbf{W}}(I_s)\|
\end{equation}
Again, $C$, $H$, and $W$ are the number of channels, height, and width of the generated image $F_{\textbf{W}}(I_s)$.  It is given a low weight overall.

\subsubsection{Total Loss.}
The total loss is comprised from a weighted sum of all above mentioned losses.
\begin{equation}
\label{eq:loss}
\mathcal{L}_{\text{total}} = \mathcal{L}_{\text{content}} + 0.4 \cdot \mathcal{L}_{\text{texture}} + 0.1 \cdot \mathcal{L}_{\text{color}} + 400 \cdot \mathcal{L}_{\text{tv}},
\end{equation}
Ignatov \textit{et al.} \cite{ignatov2017dslr} use the \textit{relu}\makebox[1mm]{\hrulefill}5\makebox[1mm]{\hrulefill}4 layer of the VGG-19 network, and mention that the above coefficients where chosen in experiments run on the DPED dataset.

\section{Experiments and Results}

\subsection{Experiments}

\begin{figure}[htb]
\includegraphics[width=\linewidth]{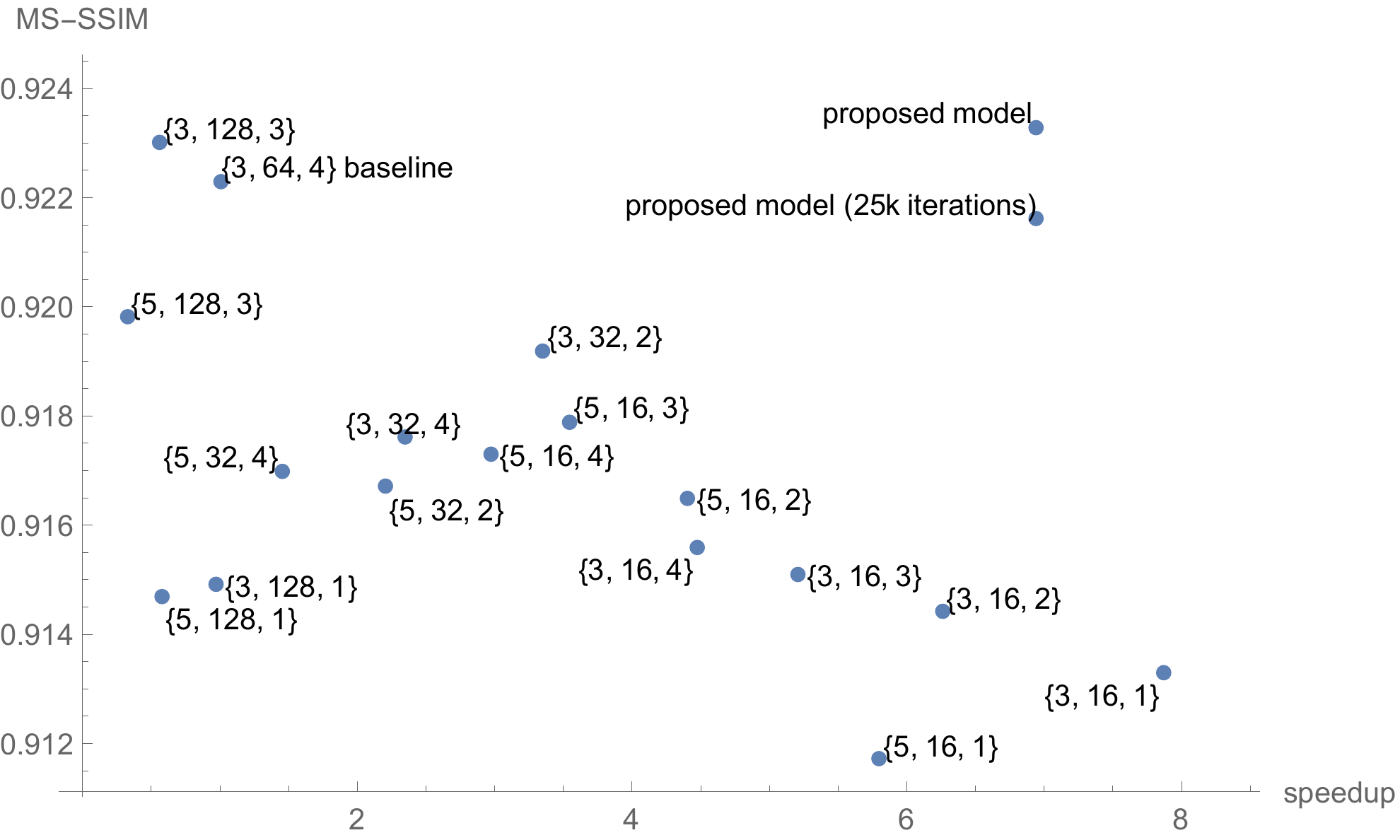}
\caption{Speedup (relative to the baseline) vs. MS-SSIM results on DPED test images, from adjusting residual CNN parameters. Key: \{kernel size, channels, blocks\}. Proposed method for reference. All models trained for 25k iterations, except for the proposed model, at 40k.}\label{fig:plot}
\end{figure}

\begin{table}[htb]

\centering
\caption{Average PSNR/SSIM results on DPED test images, using the original residual CNN architecture with adjusted parameters. 25k iterations, batch size 50.\label{tab:resnet}}
\begin{tabular}{cccccc}
\toprule
Kernel size & Channels & Blocks & Time (s) & PSNR & MS-SSIM \\ \hline
3 & 64 & 4 & 25.155 & \textbf{22.6225} & 0.9223\\
3 & 16 & 1 & \textbf{6.885} & 22.1479 & 0.9133\\
3 & 16 & 2 & 8.629 & 22.0441 & 0.9144\\
3 & 16 & 3 & 10.376 & 22.1148 & 0.9151\\
3 & 16 & 4 & 12.106 & 22.1362 & 0.9156\\
3 & 32 & 2 & 16.137 & 22.3807 & 0.9192\\
3 & 32 & 4 & 23.106 & 22.3300 & 0.9176\\
3 & 128 & 1 & 59.775 & 22.4285 & 0.9149\\
3 & 128 & 3 & 95.532 & 22.2768 & \textbf{0.9230}\\
5 & 16 & 1 & 9.297 & 21.8157 & 0.9117\\
5 & 16 & 2 & 12.332 & 21.6677 & 0.9165\\
5 & 16 & 3 & 15.211 & 22.0704 & 0.9179\\
5 & 16 & 4 & 18.243 & 21.9391 & 0.9173\\
5 & 32 & 2 & 24.538 & 21.9434 & 0.9167\\
5 & 32 & 4 & 37.137 & 21.5100 & 0.9170\\
5 & 128 & 1 & 93.066 & 22.0770 & 0.9147\\
5 & 128 & 3 & 164.068 & 21.5695 & 0.9198\\
\bottomrule
\end{tabular}
\end{table}

\subsubsection{Adjusting Residual CNN Parameters.}
\label{seq:resnet}
In order to gain an understanding of the performance properties of the DPED model~\cite{ignatov2017dslr}, the baseline's residual CNN was modified in the number of filters (or channels) each layer would have, the size of each filter's kernel, and the number of residual blocks there would be in total. While reducing the number of blocks was effective and increasing the performance, and decreasing the number of features even more so, this came at a large cost in image quality. Kernel sizes of $5\times5$ were also attempted instead of $3\times3$, but did not provide the quality improvements necessary to justify their computational costs.

In Fig.~\ref{fig:plot} and Table~\ref{tab:resnet}, a frontier can be seen, beyond which this simple architecture tuning cannot reach. More sophisticated improvements must therefore be explored. 

\subsubsection{Parametric ReLU.}
Parametric ReLU~\cite{he2015delving} is an activation function defined as
\begin{equation}
\text{PReLU}\left( y _ { i } \right) = \left\{ \begin{array} { l l } { y _ { i } , } & { \text { if } y _ { i } > 0 } \\ { a _ { i } y _ { i } , } & { \text { if } y _ { i } \leq 0 } \end{array} \right.
\end{equation}
where $y _ { i }$ is the $i$-th element of the feature vector, and $a _ { i }$ is the $i$-th element of the PReLU learned parameter vector. This permits the network to learn a slope for the ReLU activation function instead of leaving it at a constant 0 for negative inputs. In theory, this would cause the network to learn faster, prevent ReLUs from going dormant, and overall provide more power for the network at a small performance cost. 

In practice though (see an example in Table~\ref{tab:scores}), this cost was more than what was hoped, and it did not perceptibly increase the image quality.

\begin{figure}[htbp]
\includegraphics[width=\linewidth]{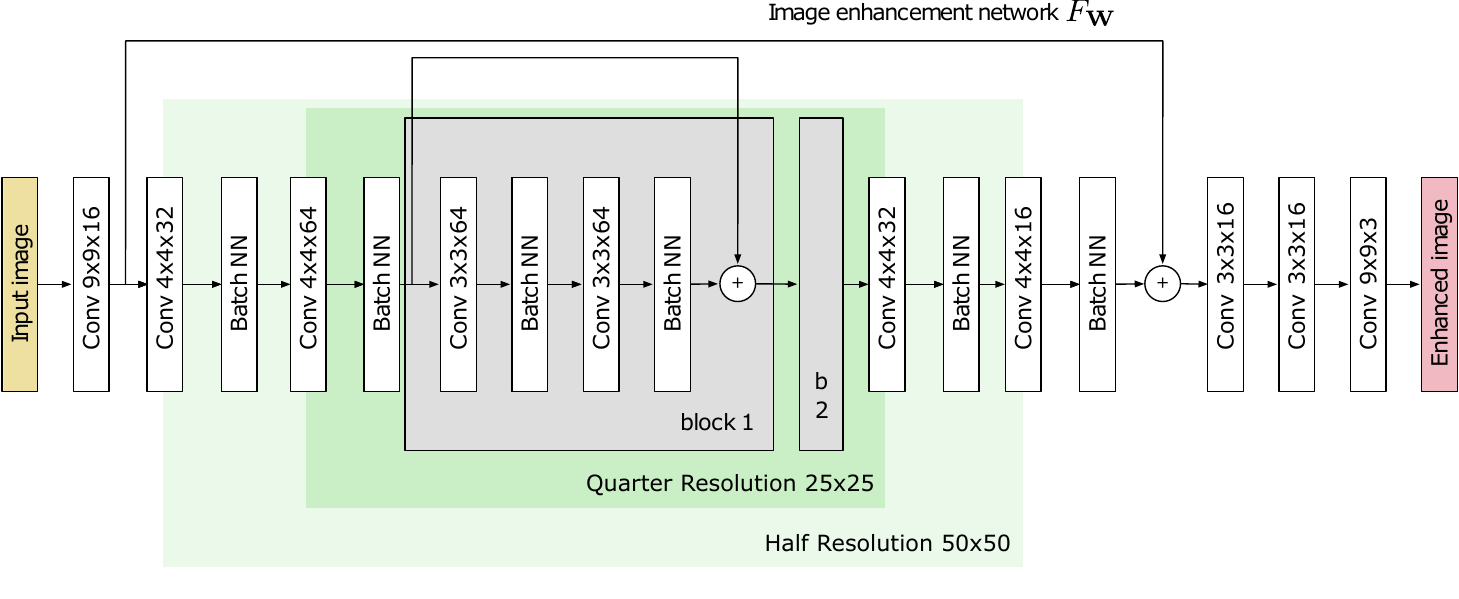}
\caption{The generator architecture of the proposed method. Discriminator and losses are the same as in the baseline.}
\label{fig:proposed_architecture}
\end{figure}

\subsubsection{Strided and Transposed Convolutions.}
In order to more drastically reduce the computation time requirements, a change in the original architecture was implemented, where the spatial resolution of the feature maps is halved, and subsequently halved again, using strided convolutional layers. At the same time, each of these strided layers doubles the number of feature maps, as suggested by Johnson~\textit{et al.}~\cite{johnson2016perceptual}.

This down-sampling operation is followed by two residual blocks at this new, $4\times$ reduced resolution, which is then followed by transposed (fractionally strided) convolution layers, which scale the feature map back up to its original resolution, using a trainable up-sampling convolution.

At each resolution, the previous feature maps of the same resolution are added to the new maps, through skip connections, in order to facilitate this network to learn simple, non-destructive transformations like the identity function.

This new architecture introduced slight checkerboard artifacts related to the upscaling process, but overall, it allowed for a much faster model without the loss in quality associated with the more straightforward approaches previously described. In Table~\ref{tab:scores} are summarized the quantitative results for several configurations.

\subsection{Results}

\begin{figure}[htbp!]

\setlength{\tabcolsep}{1pt}
\begin{tabular}{cccccccccccc}
\multicolumn{3}{c}{iPhone 3GS} & \multicolumn{3}{c}{Baseline~\cite{ignatov2017dslr}} & \multicolumn{3}{c}{\textbf{Ours}} & \multicolumn{3}{c}{DSLR Original}\\
   \multicolumn{3}{c}{\includegraphics[width=0.24556\linewidth]{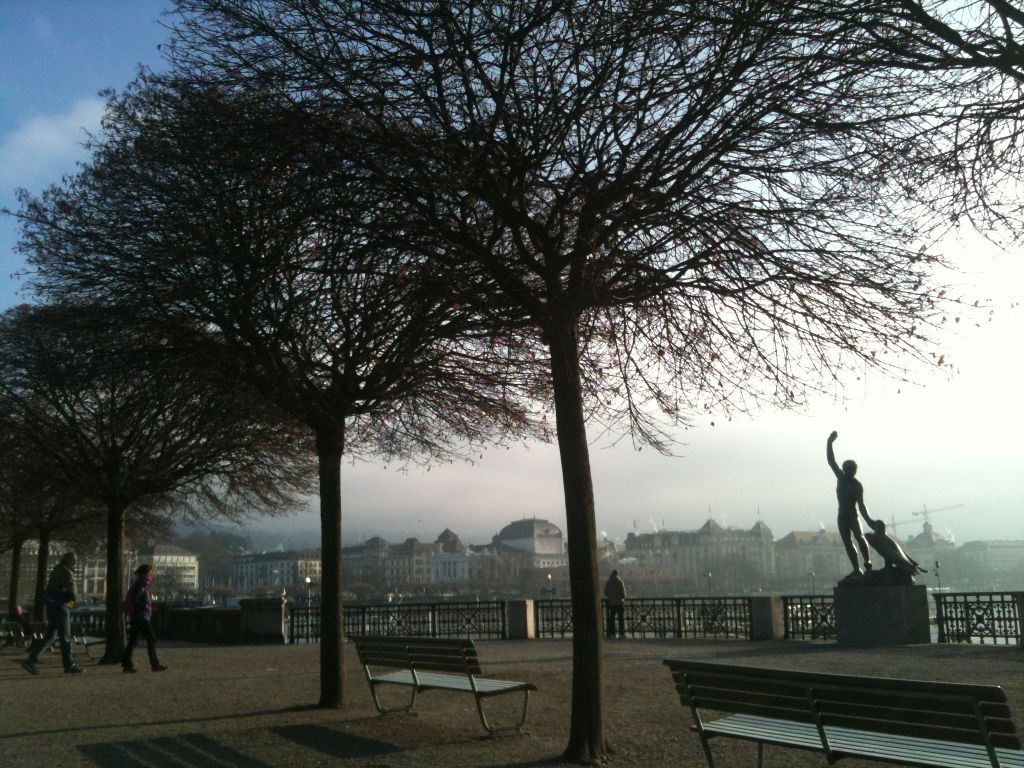}}&
   \multicolumn{3}{c}{\includegraphics[width=0.24556\linewidth]{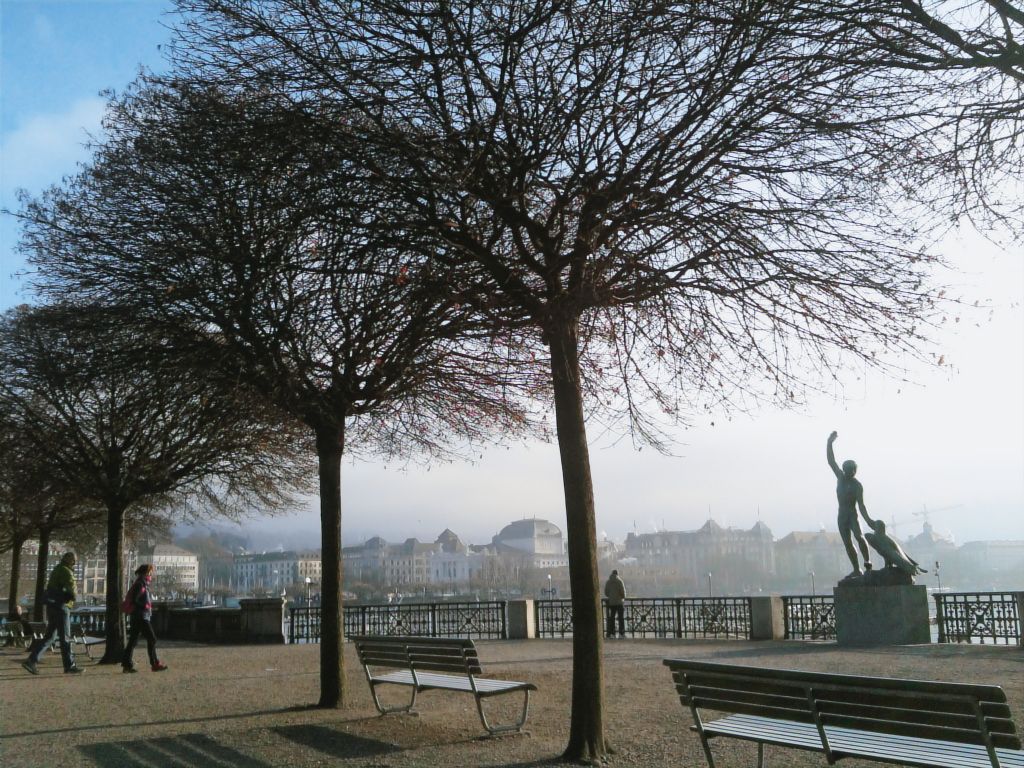}}&
   \multicolumn{3}{c}{\includegraphics[width=0.24556\linewidth]{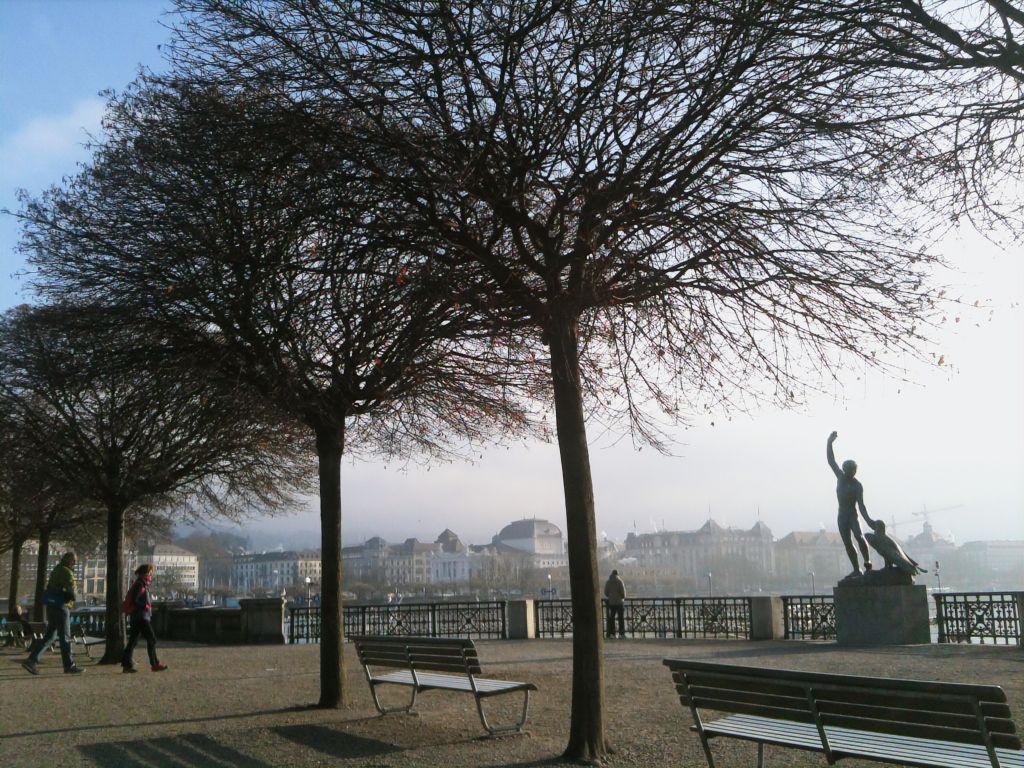}}&
   \multicolumn{3}{c}{\includegraphics[width=0.24556\linewidth]{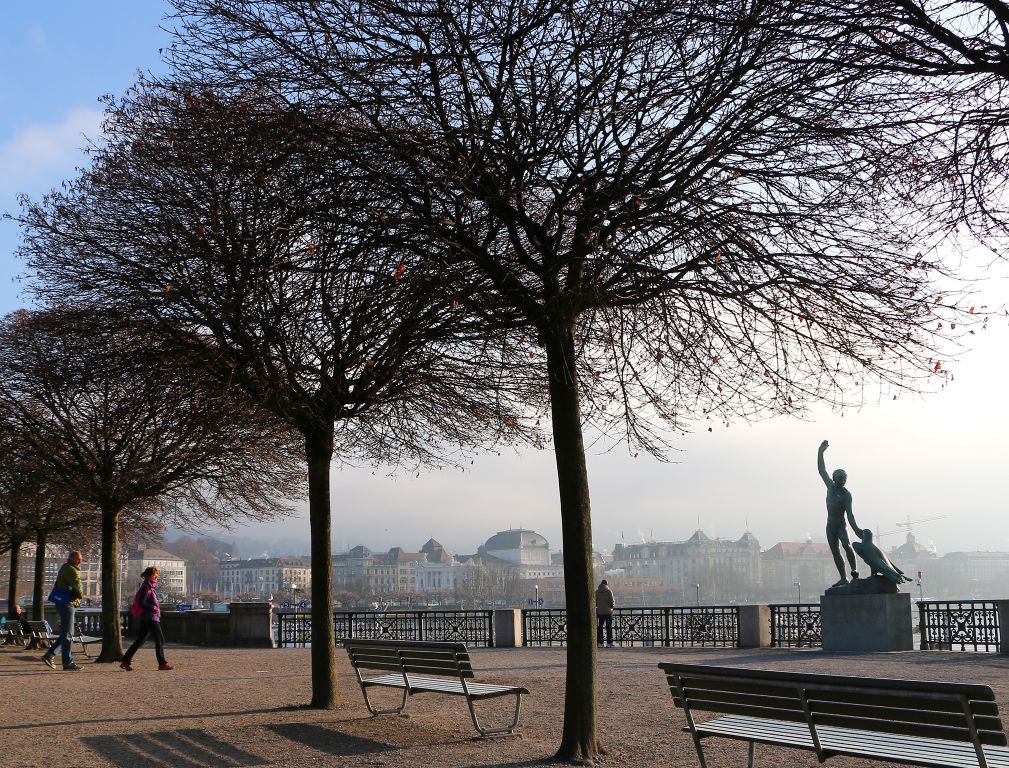}}\\
   
      \includegraphics[width=0.0779\linewidth]{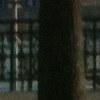}&   \includegraphics[width=0.0779\linewidth]{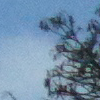}&   \includegraphics[width=0.0779\linewidth]{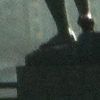}&
   \includegraphics[width=0.0779\linewidth]{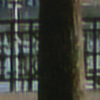}&   \includegraphics[width=0.0779\linewidth]{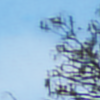}&   \includegraphics[width=0.0779\linewidth]{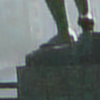}&
   \includegraphics[width=0.0779\linewidth]{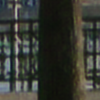}&   \includegraphics[width=0.0779\linewidth]{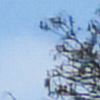}&   \includegraphics[width=0.0779\linewidth]{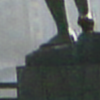}&
   \includegraphics[width=0.0779\linewidth]{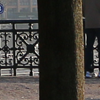}&   \includegraphics[width=0.0779\linewidth]{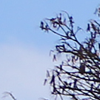}&   \includegraphics[width=0.0779\linewidth]{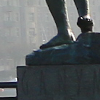}\\
   
      \multicolumn{3}{c}{\includegraphics[width=0.24556\linewidth]{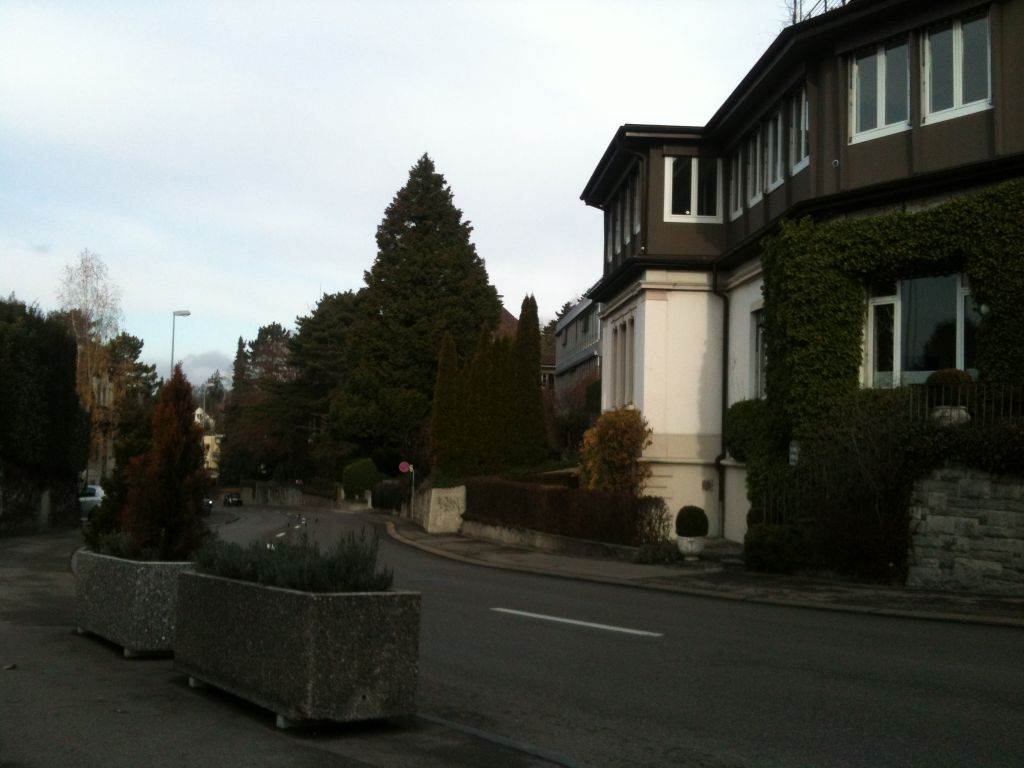}}&
   \multicolumn{3}{c}{\includegraphics[width=0.24556\linewidth]{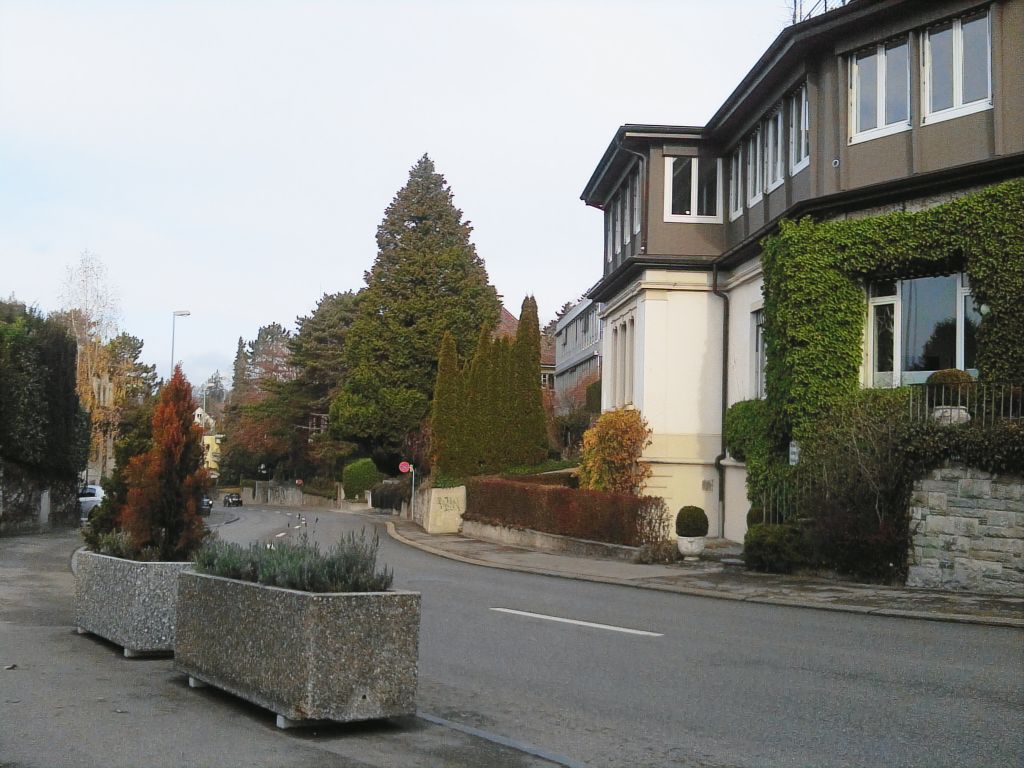}}&
   \multicolumn{3}{c}{\includegraphics[width=0.24556\linewidth]{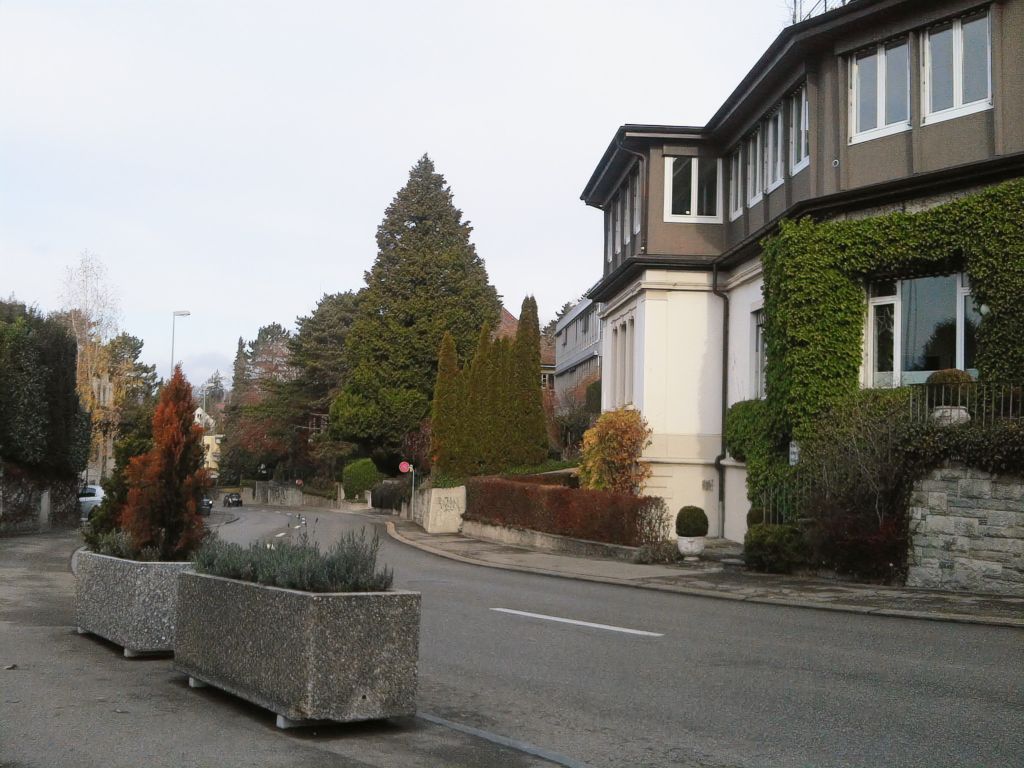}}&
   \multicolumn{3}{c}{\includegraphics[width=0.24556\linewidth]{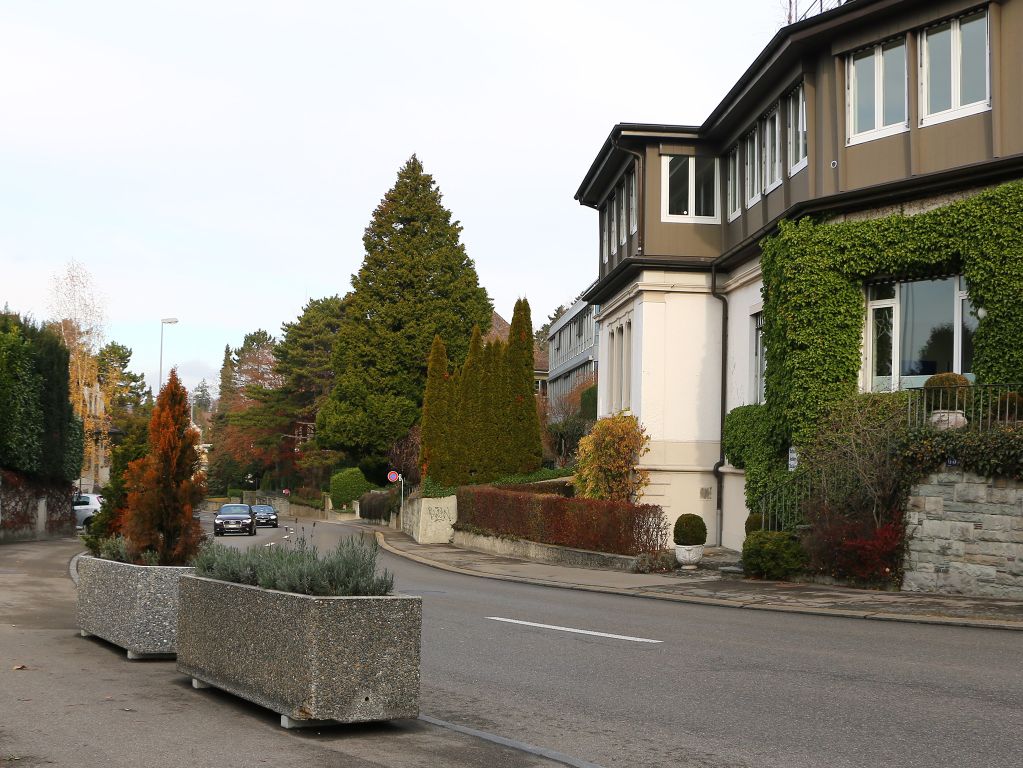}}\\
   
     \includegraphics[width=0.0779\linewidth]{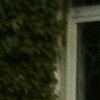}&      \includegraphics[width=0.0779\linewidth]{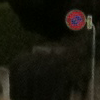}&      \includegraphics[width=0.0779\linewidth]{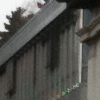}&
   \includegraphics[width=0.0779\linewidth]{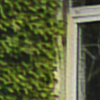}&   \includegraphics[width=0.0779\linewidth]{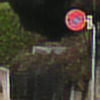}&   \includegraphics[width=0.0779\linewidth]{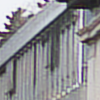}&
   \includegraphics[width=0.0779\linewidth]{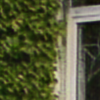}&   \includegraphics[width=0.0779\linewidth]{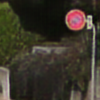}&   \includegraphics[width=0.0779\linewidth]{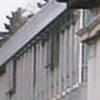}&
   \includegraphics[width=0.0779\linewidth]{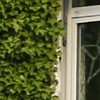}&   \includegraphics[width=0.0779\linewidth]{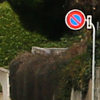}&   \includegraphics[width=0.0779\linewidth]{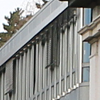}\\
   
      \multicolumn{3}{c}{\includegraphics[width=0.24556\linewidth]{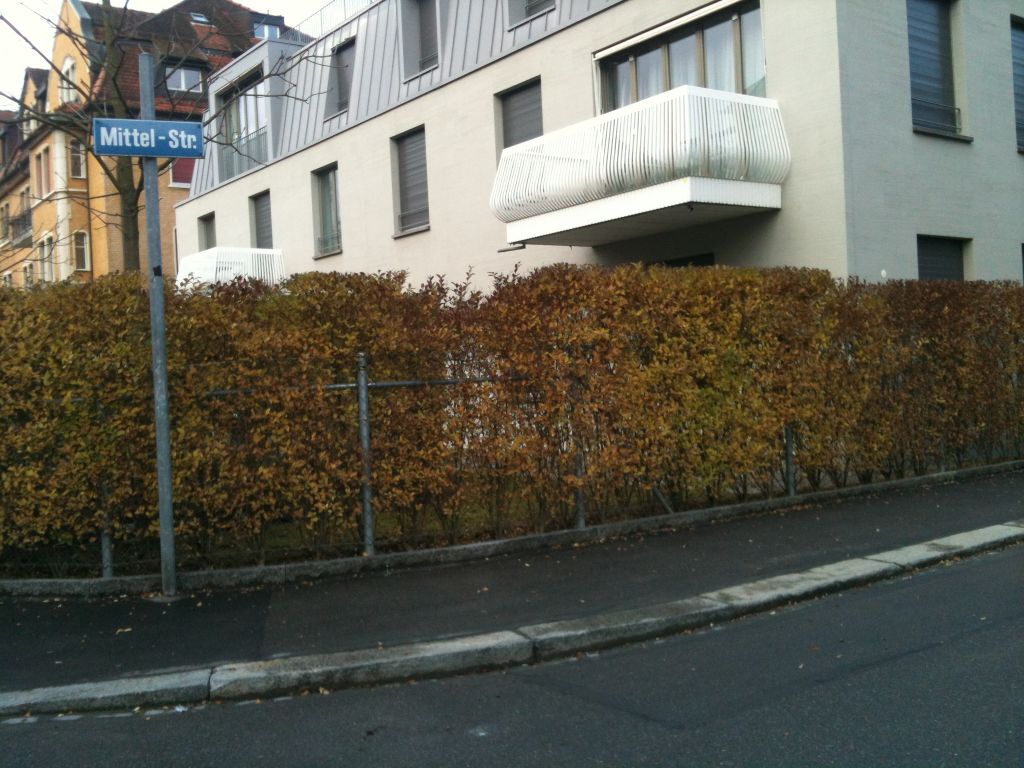}}&
   \multicolumn{3}{c}{\includegraphics[width=0.24556\linewidth]{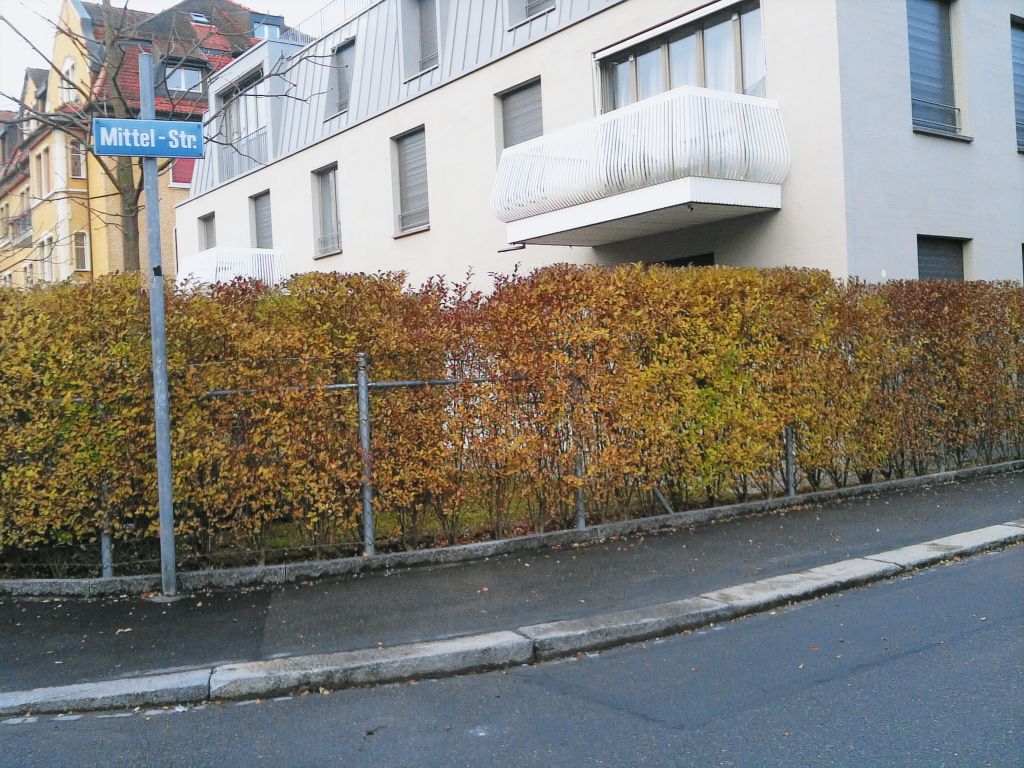}}&
   \multicolumn{3}{c}{\includegraphics[width=0.24556\linewidth]{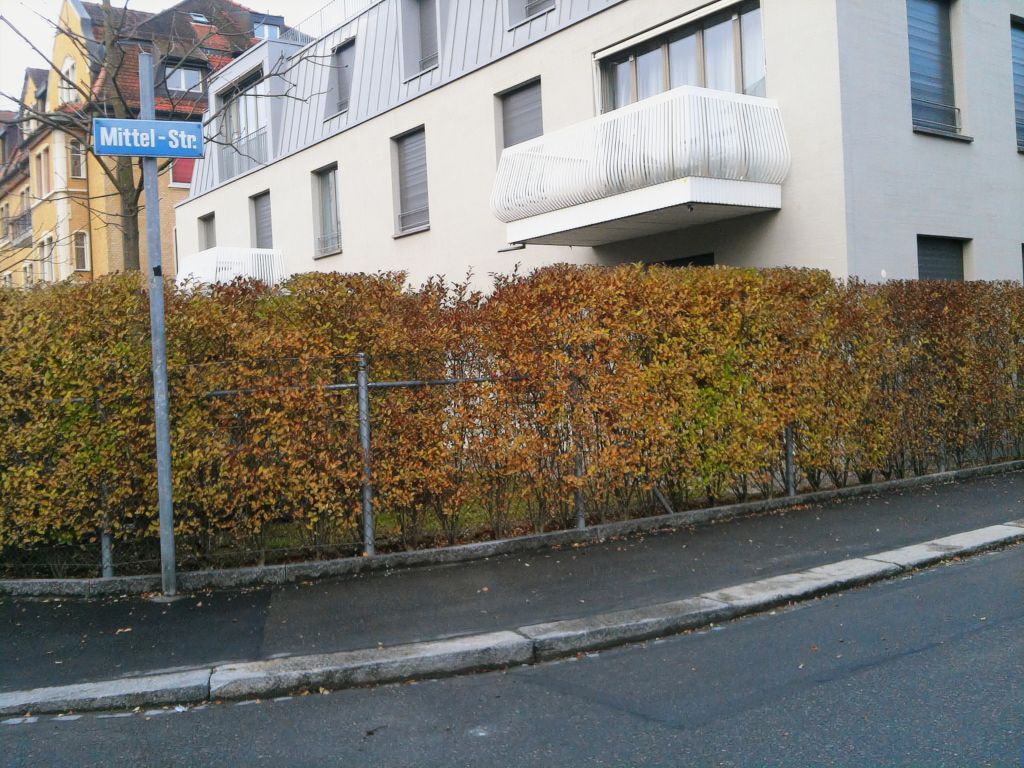}}&
   \multicolumn{3}{c}{\includegraphics[width=0.24556\linewidth]{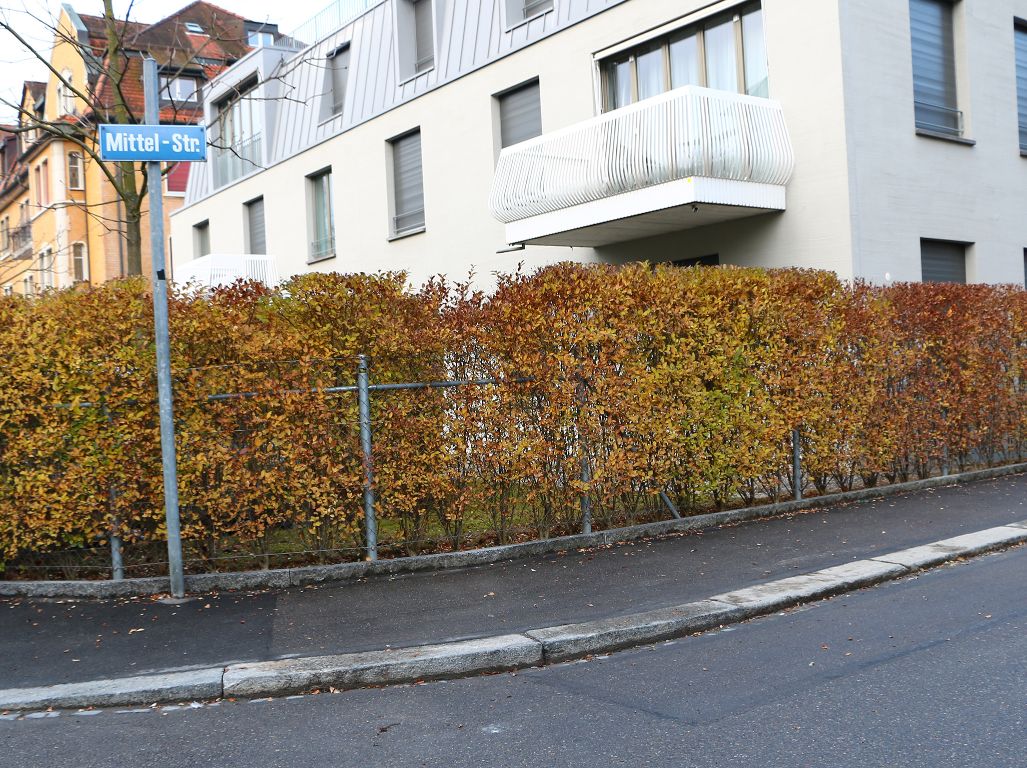}}\\
   
         \includegraphics[width=0.0779\linewidth]{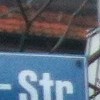}&      \includegraphics[width=0.0779\linewidth]{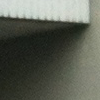}&      \includegraphics[width=0.0779\linewidth]{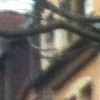}&
   \includegraphics[width=0.0779\linewidth]{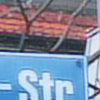}&   \includegraphics[width=0.0779\linewidth]{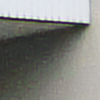}&   \includegraphics[width=0.0779\linewidth]{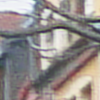}&
   \includegraphics[width=0.0779\linewidth]{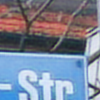}&   \includegraphics[width=0.0779\linewidth]{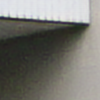}&   \includegraphics[width=0.0779\linewidth]{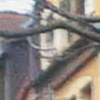}&
   \includegraphics[width=0.0779\linewidth]{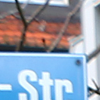}&   \includegraphics[width=0.0779\linewidth]{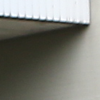}&   \includegraphics[width=0.0779\linewidth]{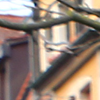}\\
   
\end{tabular}
\caption{Visual assessment. From left to right: The input test image from the iPhone 3GS, the output from the baseline model, the output from our model, and the (cropped) ground truth photograph from the DSLR camera.}
\label{fig:visual}
\end{figure}

\begin{table}[htb]

\centering
\caption{Average PSNR/SSIM results on DPED test images, using the proposed strided architecture with varying parameters. The best configuration we propose, line 3, was chosen as a compromise between quality and speed.
\label{tab:scores}}
\begin{tabular}{cccccc}
\toprule
Kernel size & Channels & PReLU & Time (s) & PSNR & MS-SSIM \\ \hline
3 & 16-64 & no & \textbf{7.729} & 22.3049 & 0.9176\\
3 & 32-128 & no & 14.641 & 22.3909 & 0.9235\\
\textit{3-4} & \textit{16-64} & \textit{no} & \textit{7.987} & \textit{22.5248} & \textit{0.9233}\\
3-4 & 32-128 & no & 15.413 & \textbf{22.6636} & \textbf{0.9248}\\
4 & 16-64 & no & 8.84 & 22.2421 & 0.9232\\
4 & 32-128 & no & 17.826 & 22.2812 & 0.9206\\
4 & 32-128 & yes & 20.584 & 22.3166 & 0.9209\\
\bottomrule
\end{tabular}
\end{table}

The best result we achieved was with this new strided approach. The generator architecture is shown in Fig.~\ref{fig:proposed_architecture}. We chose a kernel size of $3\times3$, except in the strided convolutional layers, where we opted for $4\times4$ instead, in order to mitigate the checkerboard artifacts. The number of feature maps starts at 16 and increases up to 64 in the middle of the network. We trained the network for 40k iterations using an Adam optimizer and a batch size of 50.

Our network~\footnote{Codes and models publicly released at: \url{https://github.com/dojure/FPIE/}} takes only $\SI{3.2}\second$ of CPU time to enhance a $1280\times720\mskip\thinmuskip$px image compared to the baseline's $\SI{20.5}\second$. This represents a 6.3-fold speedup. Additionally, the amount of RAM required is reduced from $\SI{3.7}{\giga\byte}$ to $\SI{2.3}{\giga\byte}$.

As part of a PIRM 2018 challenge on perceptual image enhancement on smartphones~\cite{ignatov2018pirm}, a user study was conducted where 2000 people were asked to rate the visual results (photos) of the solutions submitted by challenge participants. The users were able to rate each photo with scores of 1, 2, 3 and 4, corresponding to low and high-quality visual results. The average of all user ratings was then computed and considered as a MOS score of each solution.

\begin{table}[htb]
\centering

\caption{PIRM 2018 challenge final ranking of teams and baselines~\cite{ignatov2018pirm} \label{tab:contest}}
\setlength{\tabcolsep}{2pt}
\begin{tabular}{l|cccrrc}
\toprule
Team & PSNR & MS-SSIM & \textbf{MOS} & CPU (ms) & GPU (ms) & RAM (GB) \\ \hline
Mt.Phoenix & 21.99 & 0.9125 & 2.6804 & 682 & 64 & 1.4 \\
\textbf{EdS (Ours)} & 21.65 & 0.9048 & 2.6523 & 3241 & 253 & 2.3 \\
BOE-SBG & 21.99 & 0.9079 & 2.6283 & 1620 & 111 & 1.6 \\
MENet & 22.22 & 0.9086 & 2.6108 & 1461 & 138 & 1.8 \\
Rainbow & 21.85 & 0.9067 & 2.5583 & 828 & 111 & 1.6 \\
KAIST-VICLAB & 21.56 & 0.8948 & 2.5123 & 2153 & 181 & 2.3 \\
SNPR & 22.03 & 0.9042 & 2.4650 & 1448 & 81 & 1.6 \\
DPED (Baseline) & 21.38 & 0.9034 & 2.4411 & 20462 & 1517 & 3.7 \\
Geometry & 21.79 & 0.9068 & 2.4324 & 833 & 83 & 1.6 \\
IV SR+ & 21.60 & 0.8957 & 2.4309 & 1375 & 125 & 1.6 \\
SRCNN (Baseline) & 21.31 & 0.8929 & 2.2950 & 3274 & 204 & 2.6 \\
TEAM ALEX & 21.87 & 0.9036 & 2.1196 & 781 & 70 & 1.6 \\
\bottomrule
\end{tabular}
\end{table}

With a MOS of 2.6523, our submission (see Table~\ref{tab:contest}) scored significantly higher than the DPED baseline (2.4411) and was second only to the winning submission, which scored 2.6804. The submission was tested against a different test set, which partially explains its lower PSNR and MS-SSIM scores. It should be noted that the submission shares the same architecture as this paper's main result, but was trained for only 33k iterations. 

Differences between the DPED baseline and our result are somewhat subtle. Our model produces noticeably fewer colored artifacts around hard edges (e.g. Fig.~\ref{fig:visual}, first row, first zoom box), more accurate colors (e.g. the sky in first row, second box), as well as reduced noise in smooth shadows (last row, second box), and in dense foliage (middle row, first box), it produces more realistic textures than the baseline. Contrast, especially in vertical features (middle row, third box), is often less pronounced. However, this comes with the advantage of fewer grid-like artifacts. For more visual results of our method we refer the reader to the Appendix.

While these subjective evaluation methods are clearly in favor of our method, the PSNR and MS-SSIM scores comparing the generated images to the target DSLR photos are less conclusive. PSNR and MS-SSIM seem to be only weakly correlated with MOS~\cite{ignatov2018pirm}. Better perceptual quality metrics including ones requiring no reference images might be a promising component of future works.

\section{Conclusion}
\label{sec:conclusion}

Thanks to strided convolutions, a promising architecture was found in the quest for efficient photo enhancement on mobile hardware. Our model produces clear, detailed images exceeding the quality of the baseline, while only requiring $\SI{16}\percent$ as much computation time.

Even though, as evidenced by the PIRM 2018 challenge results~\cite{ignatov2018pirm}, further speed improvements will definitely be seen in future works, it is reassuring to conclude that convolutional neural network-based image enhancement can already produce high quality results with performance acceptable for mobile devices.

\section*{Acknowledgments}
This work was partly supported by ETH Zurich General Fund and a hardware (GPU) grant from NVIDIA.

\clearpage
\section*{Appendix. Results of the Proposed Method}
\label{sec:appendix}

\begin{figure}[!h]
\setlength{\tabcolsep}{1pt}
\begin{tabular}{cccc}
iPhone 3GS original & Enhanced with our method\\
   \includegraphics[width=0.497\linewidth]{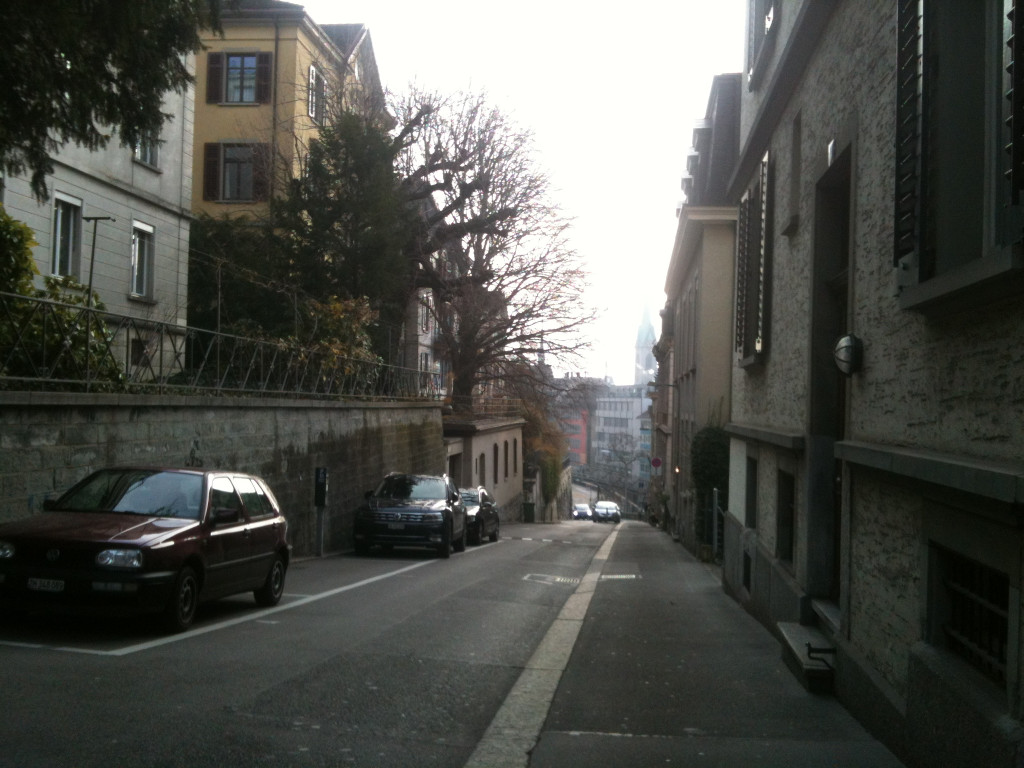}&
     \includegraphics[width=0.497\linewidth]{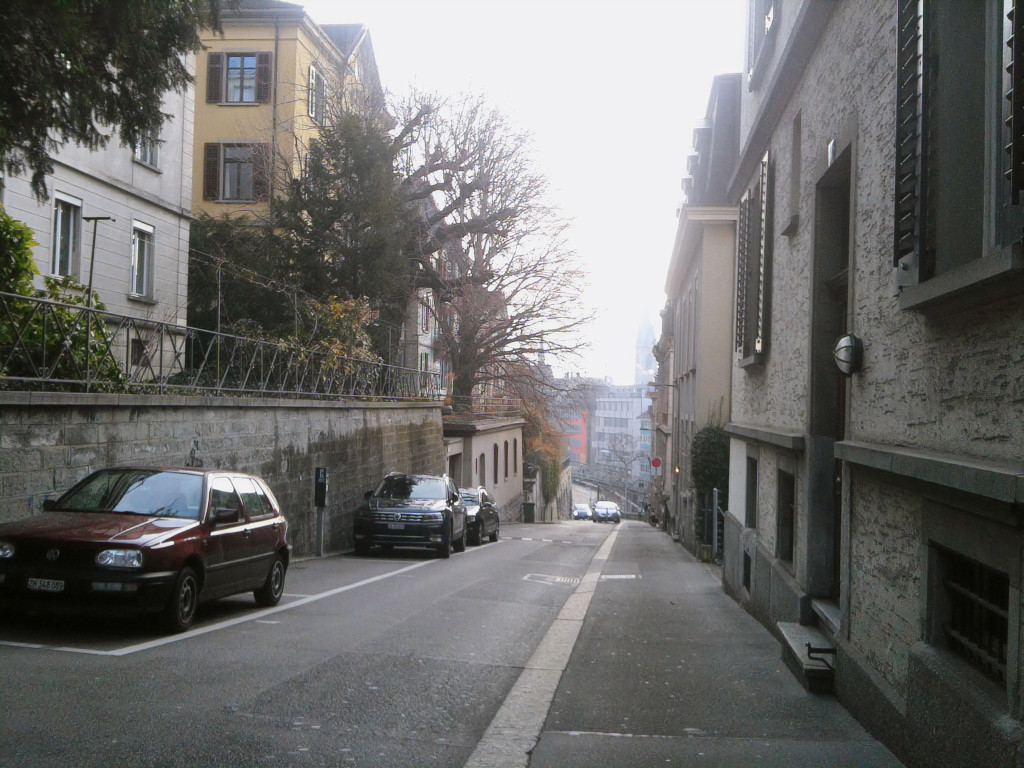}\\
   \includegraphics[width=0.497\linewidth]{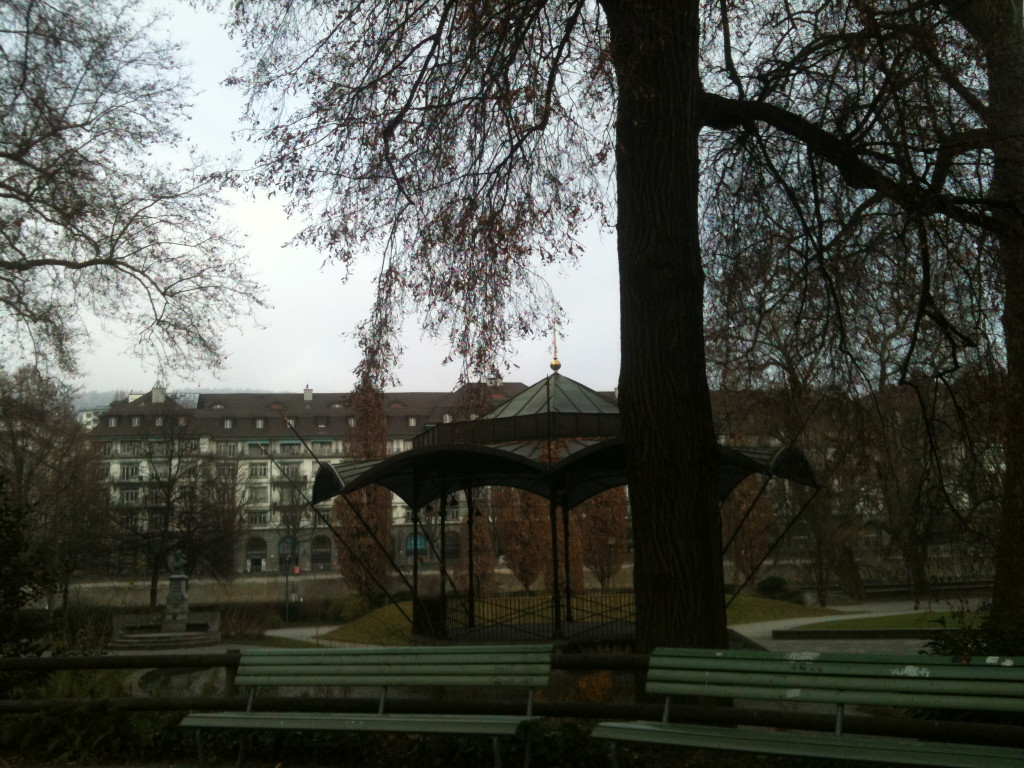}&
     \includegraphics[width=0.497\linewidth]{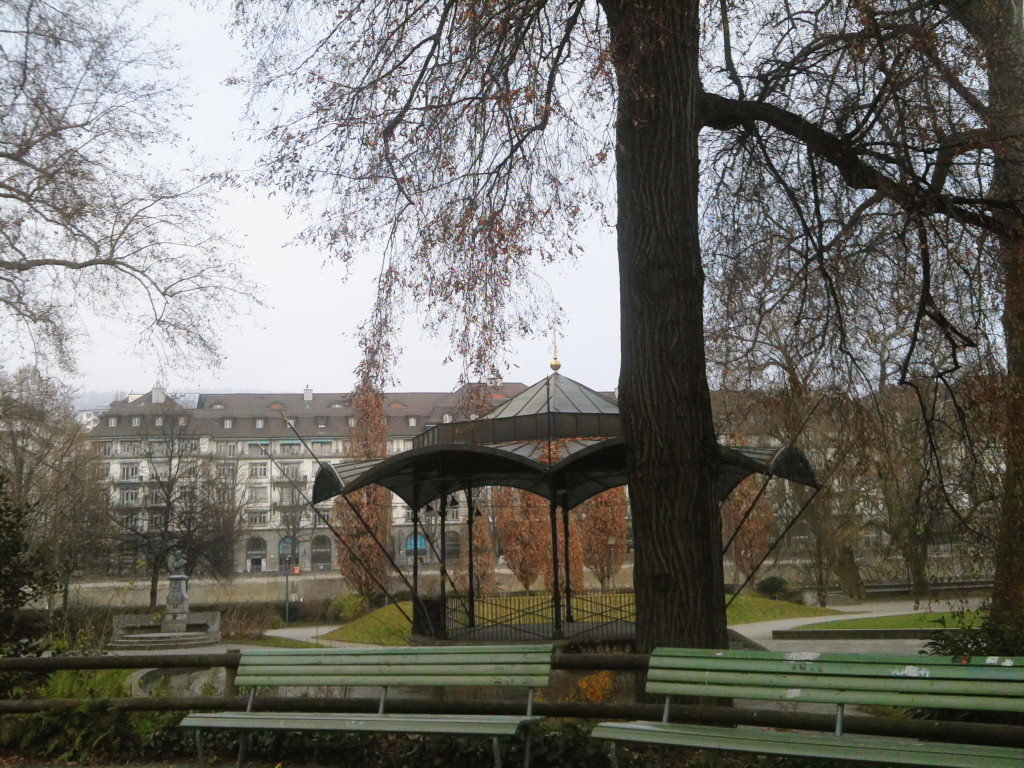}\\
   \includegraphics[width=0.497\linewidth]{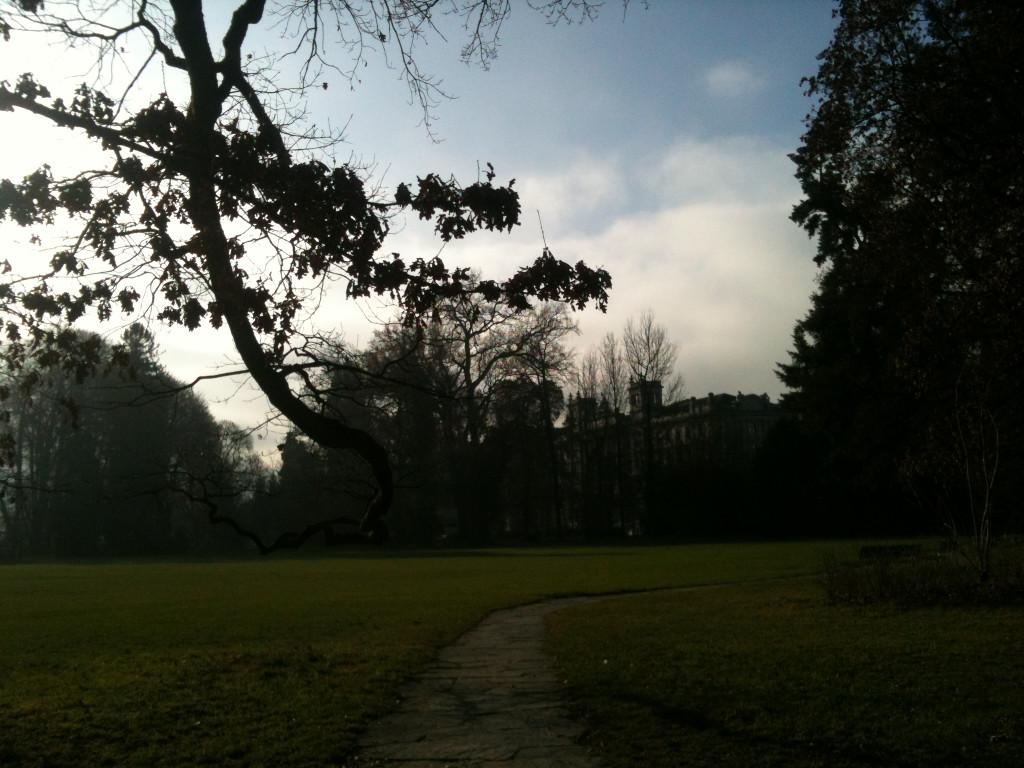}&
     \includegraphics[width=0.497\linewidth]{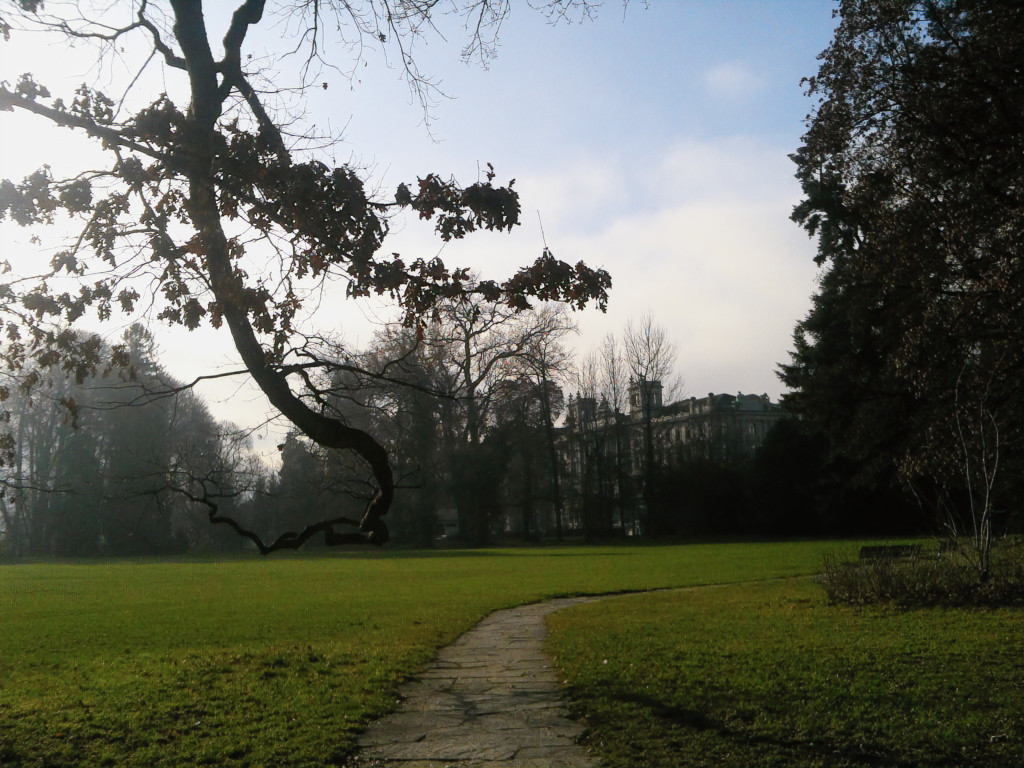}\\
\end{tabular}
\caption{Visual results for our method.}
\end{figure}

\begin{figure}[h]
\setlength{\tabcolsep}{1pt}
\begin{tabular}{cccc}
iPhone 3GS original & Enhanced with our method\\
   \includegraphics[width=0.497\linewidth]{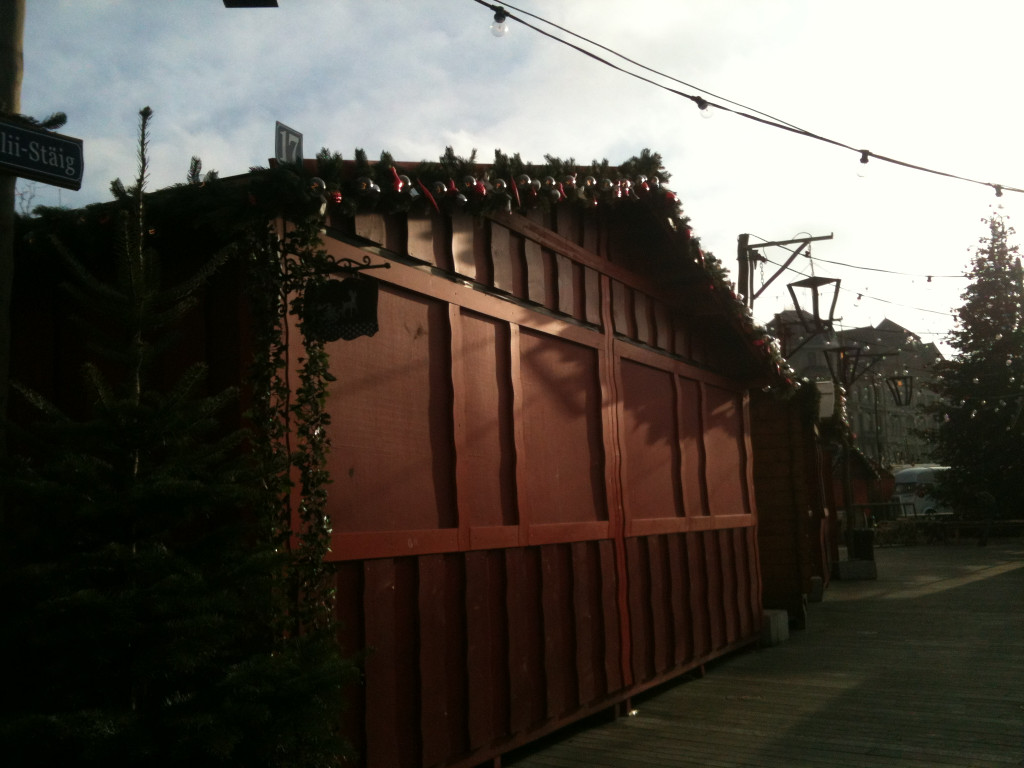}&
     \includegraphics[width=0.497\linewidth]{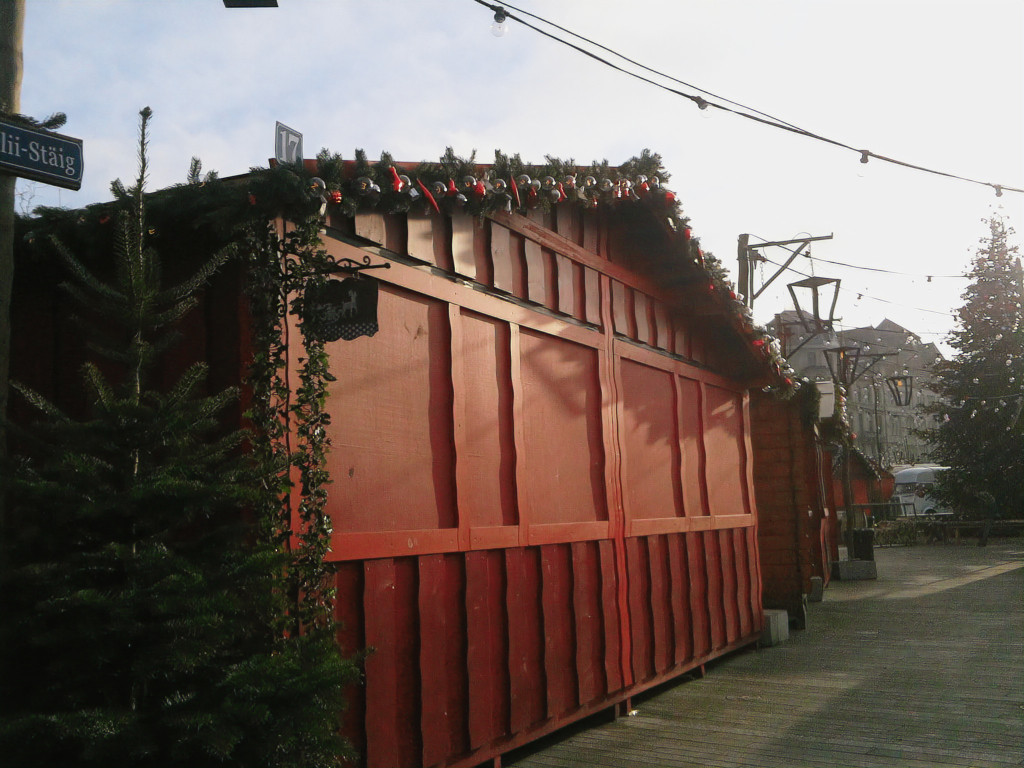}\\
   \includegraphics[width=0.497\linewidth]{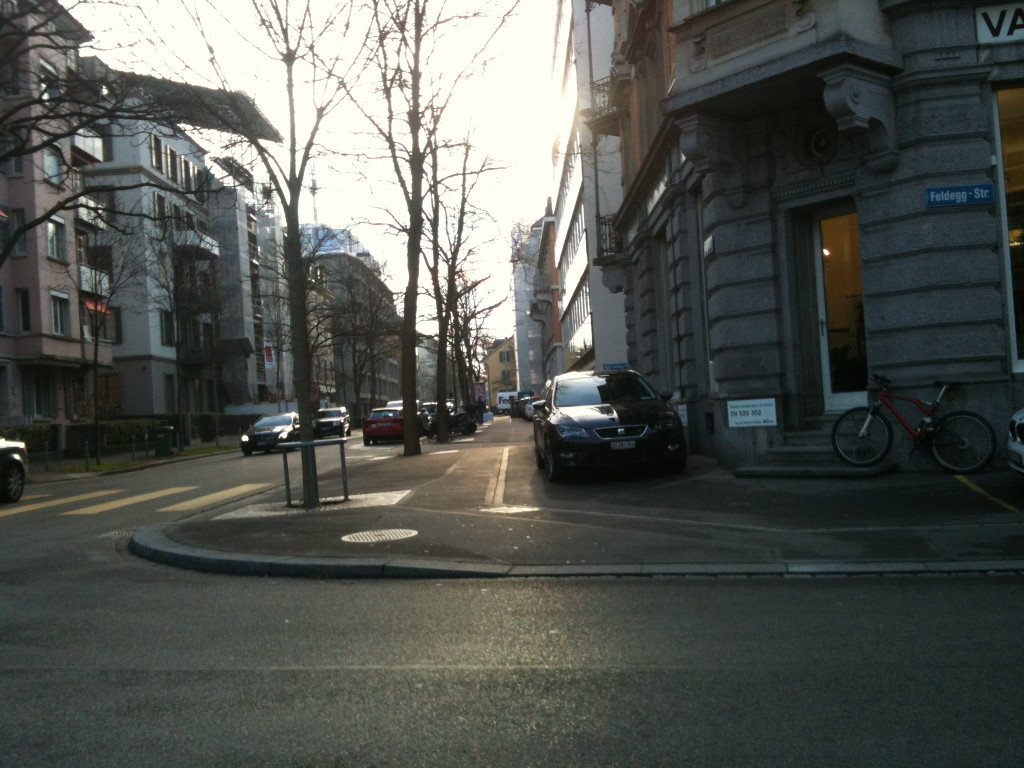}&
     \includegraphics[width=0.497\linewidth]{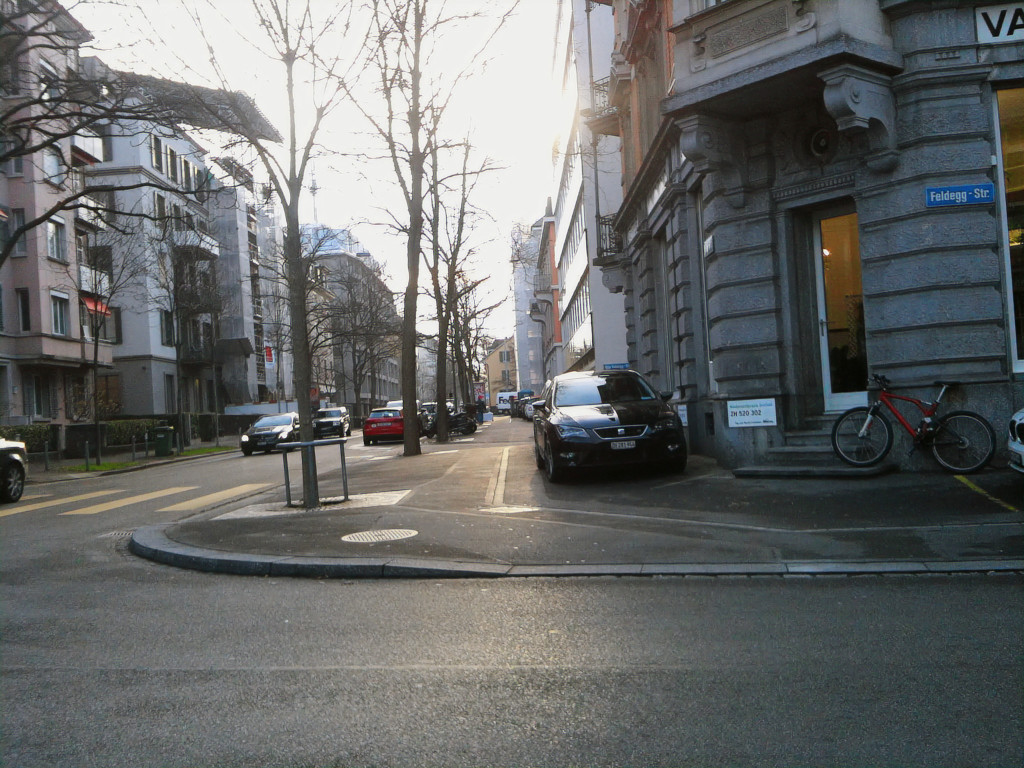}\\
   \includegraphics[width=0.497\linewidth]{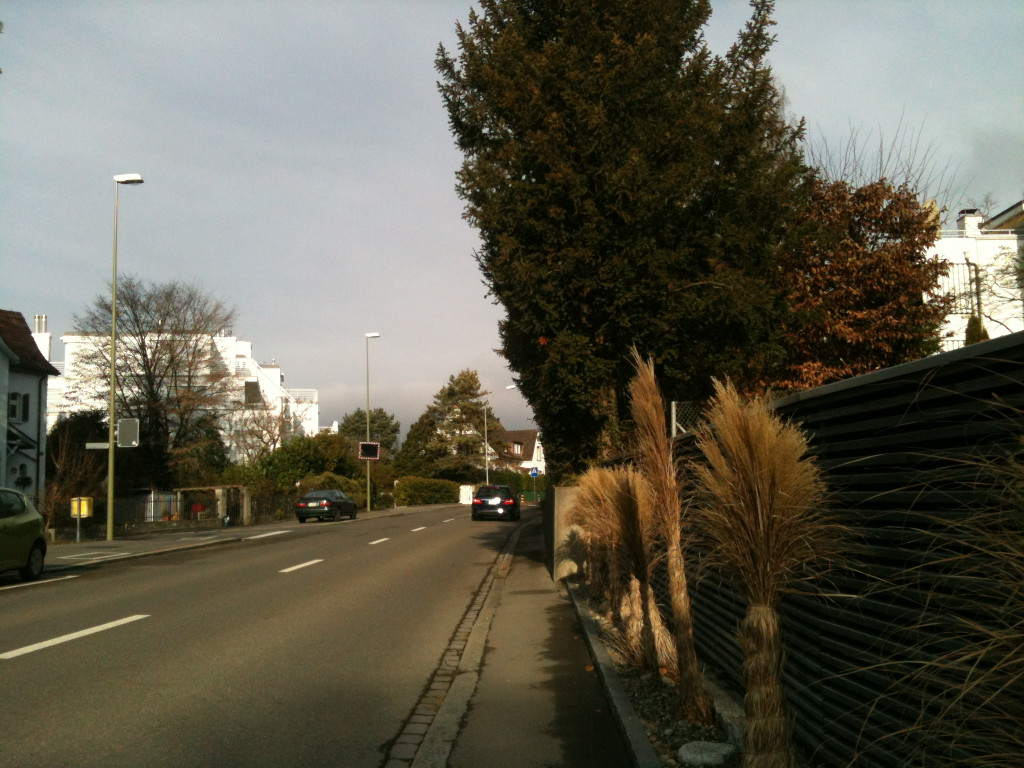}&
     \includegraphics[width=0.497\linewidth]{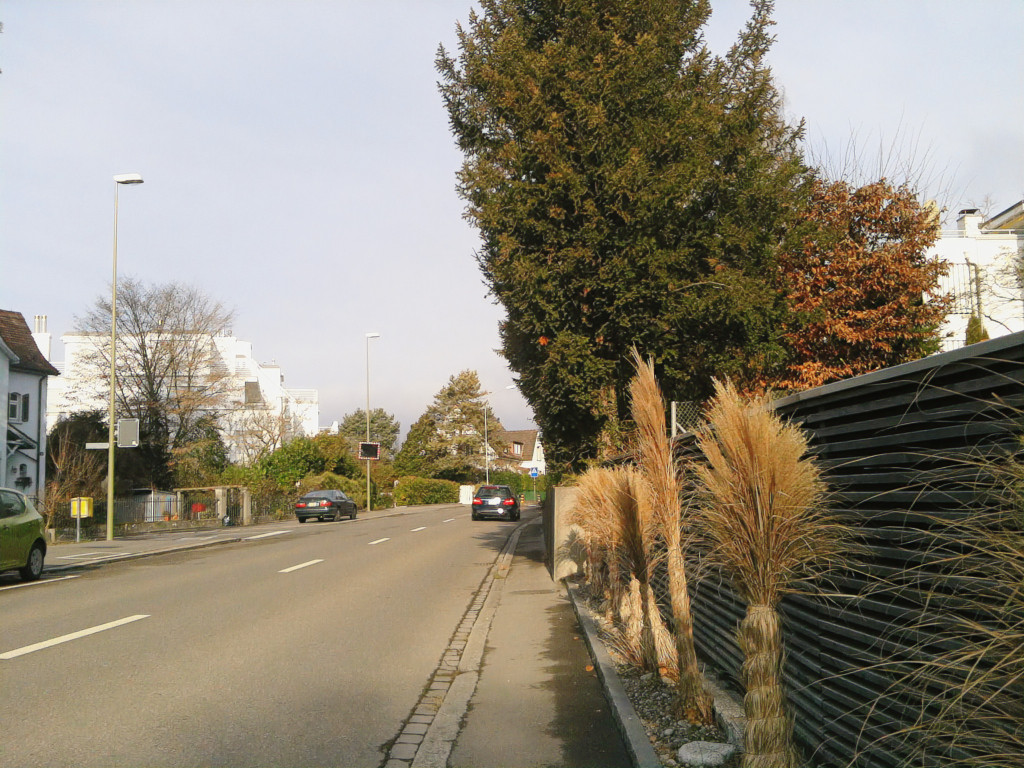}\\
\end{tabular}
\caption{Visual results for our method.}
\end{figure}

\begin{figure}[h]
\setlength{\tabcolsep}{1pt}
\begin{tabular}{cccc}
iPhone 3GS original & Enhanced with our method\\
   \includegraphics[width=0.497\linewidth]{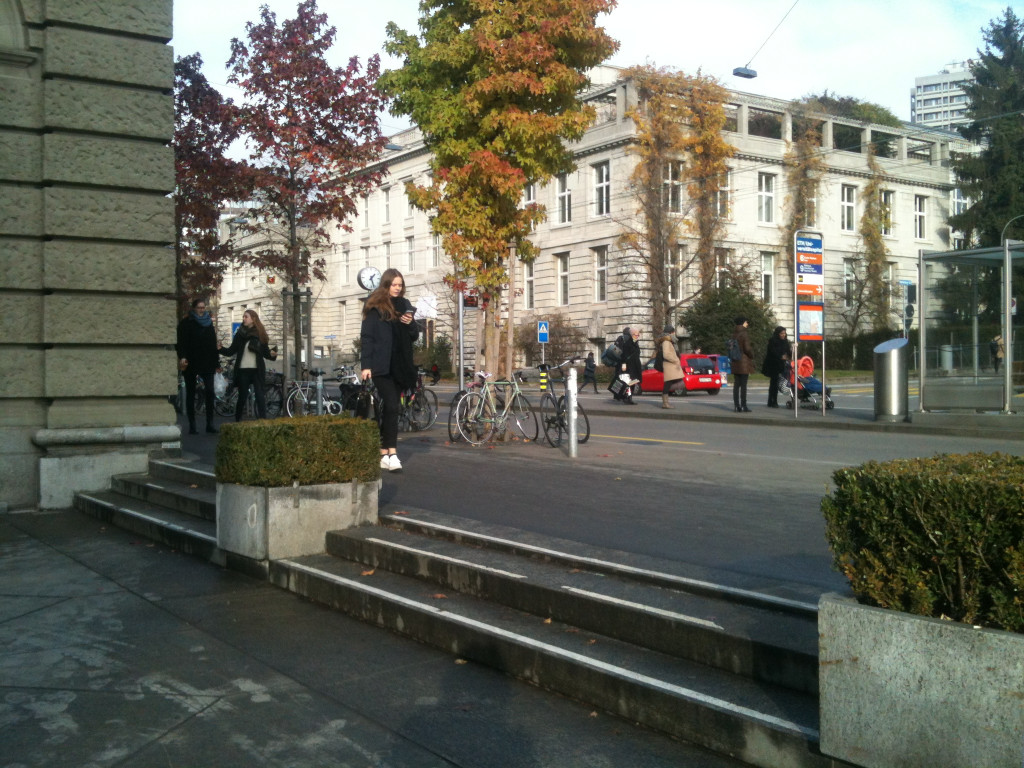}&
     \includegraphics[width=0.497\linewidth]{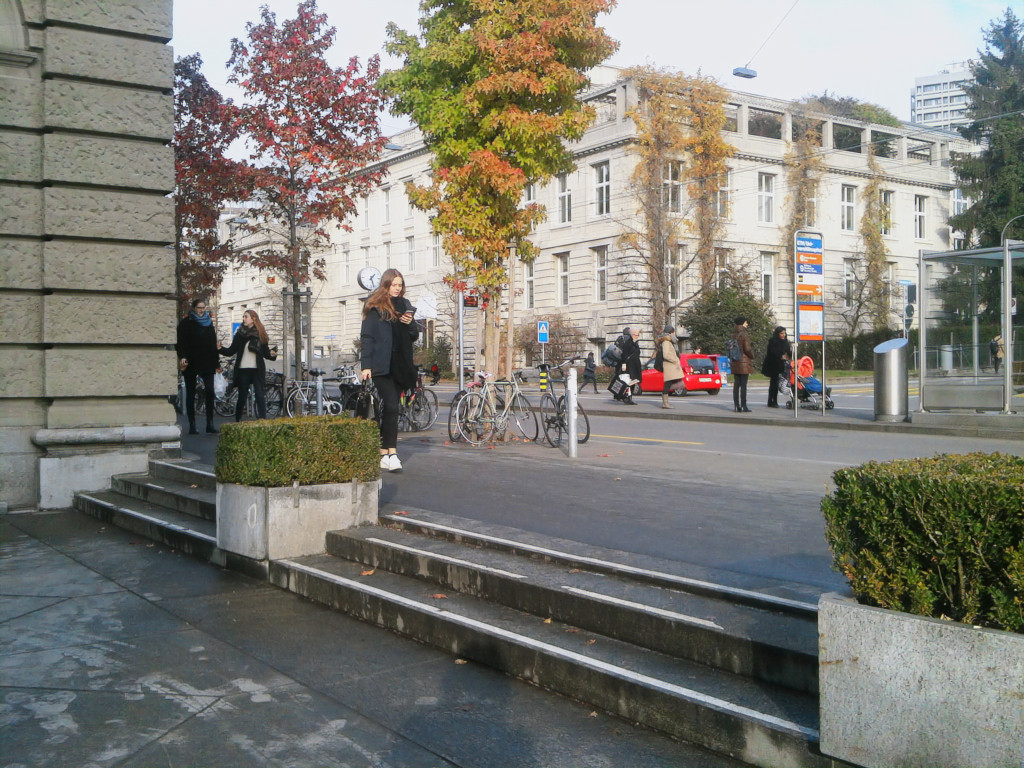}\\
   \includegraphics[width=0.497\linewidth]{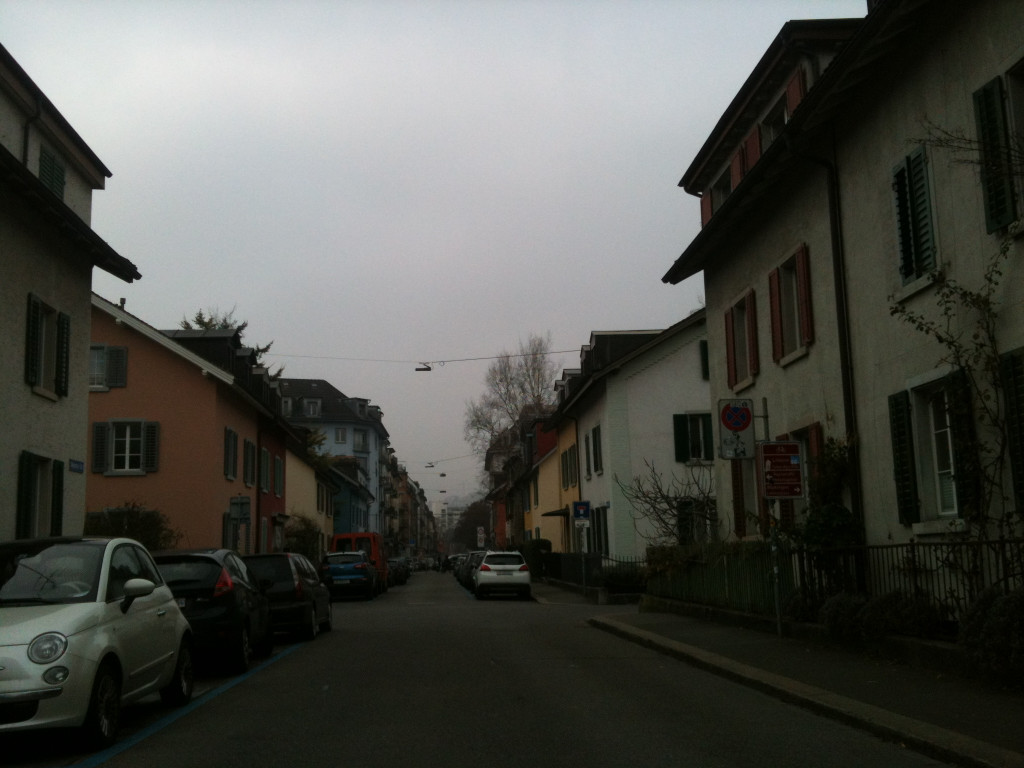}&
     \includegraphics[width=0.497\linewidth]{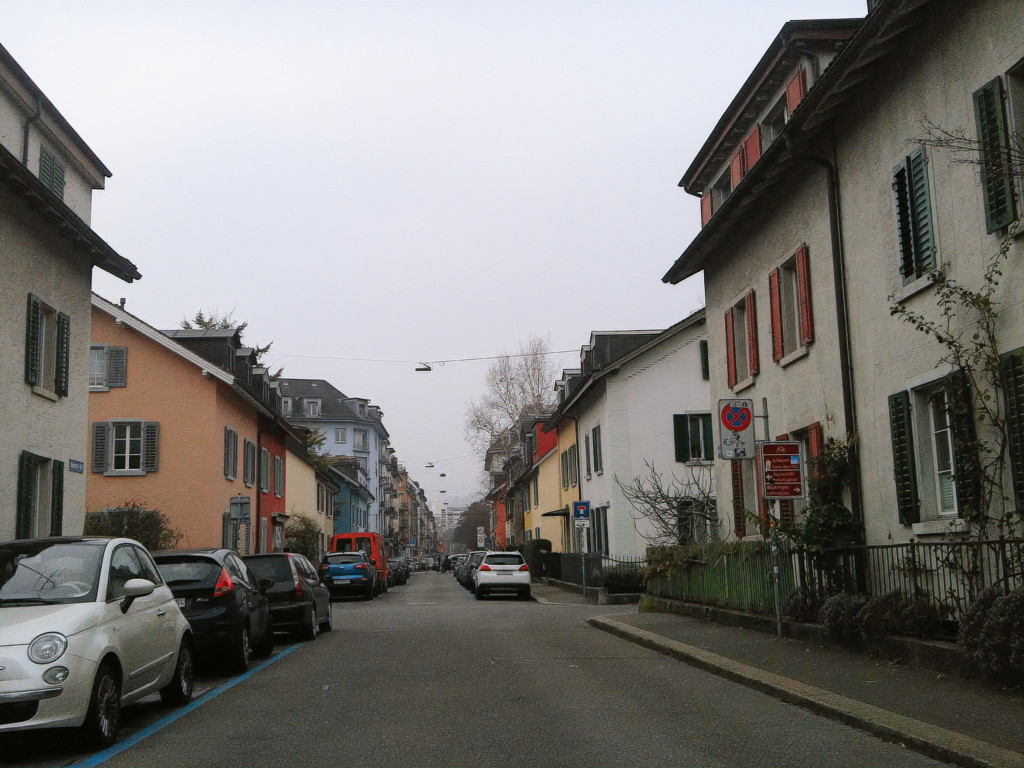}\\
   \includegraphics[width=0.497\linewidth]{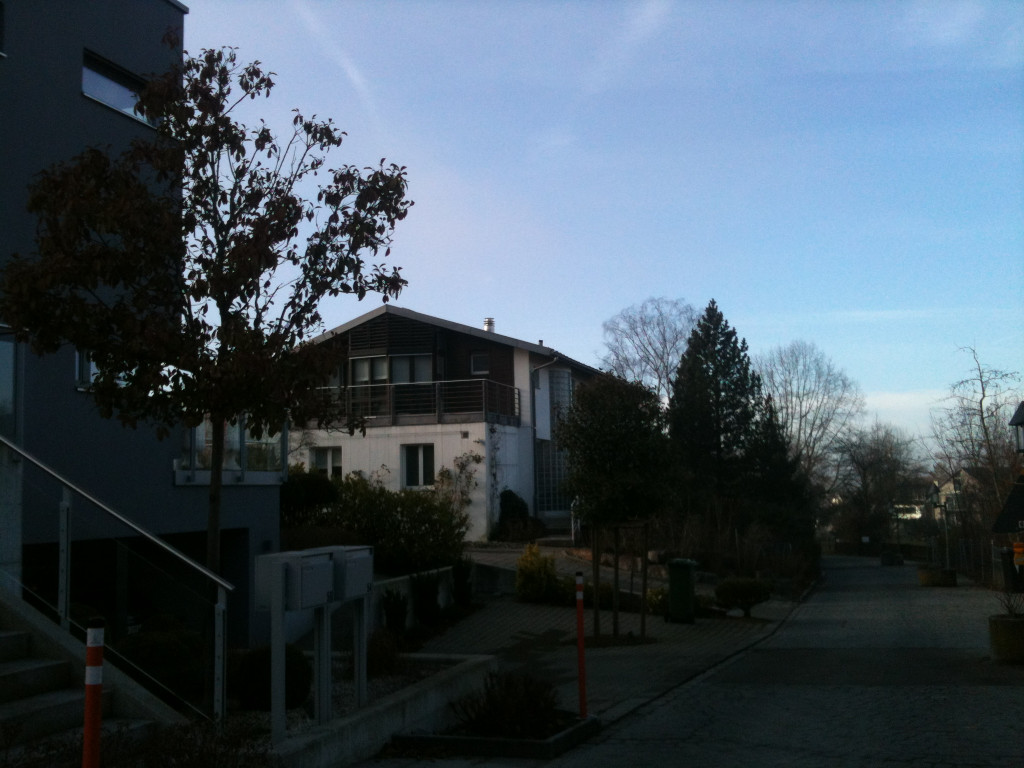}&
     \includegraphics[width=0.497\linewidth]{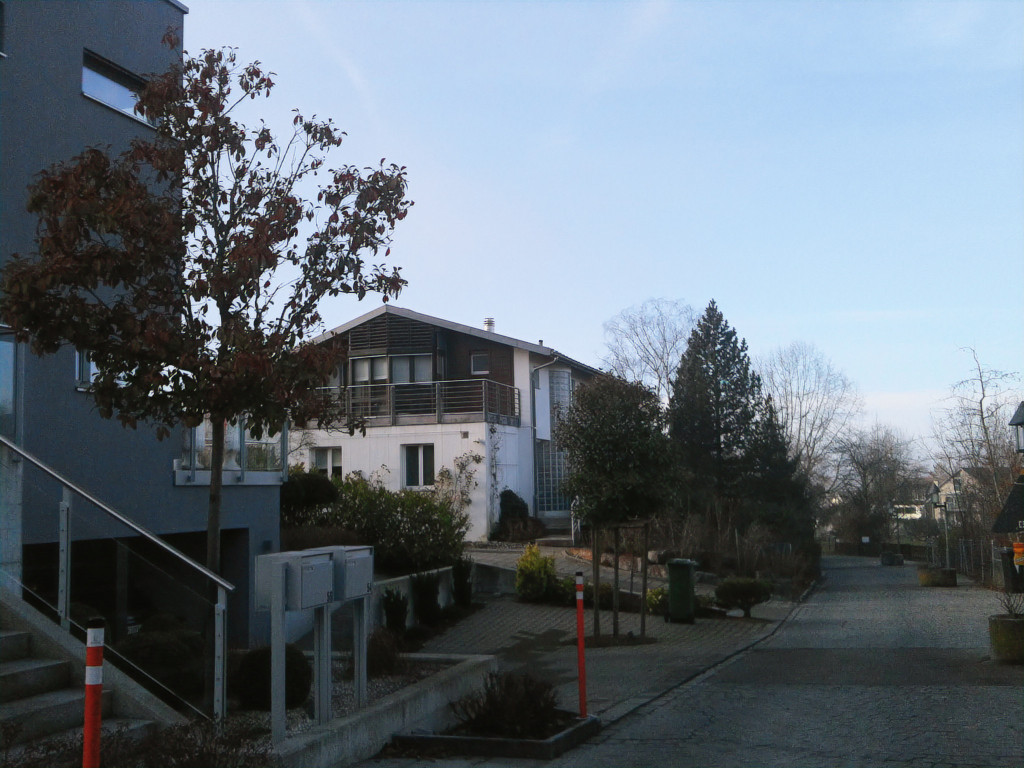}\\
\end{tabular}
\caption{Visual results for our method.}
\end{figure}

\begin{figure}[h]
\setlength{\tabcolsep}{1pt}
\begin{tabular}{cccc}
iPhone 3GS original & Enhanced with our method\\
   \includegraphics[width=0.497\linewidth]{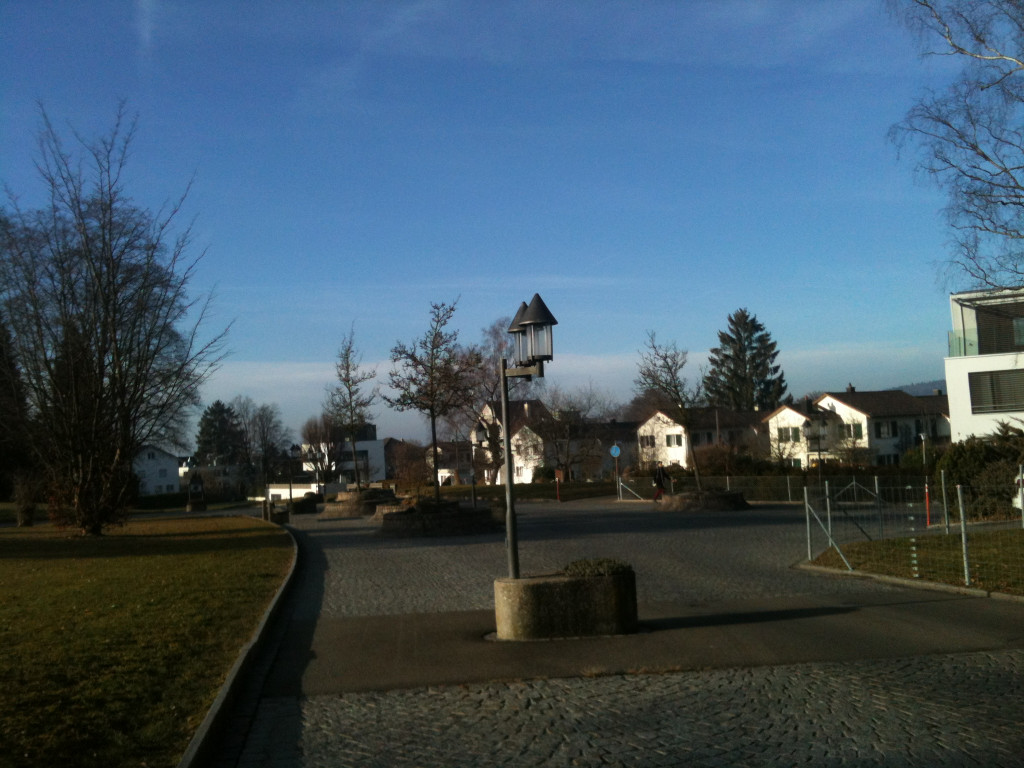}&
     \includegraphics[width=0.497\linewidth]{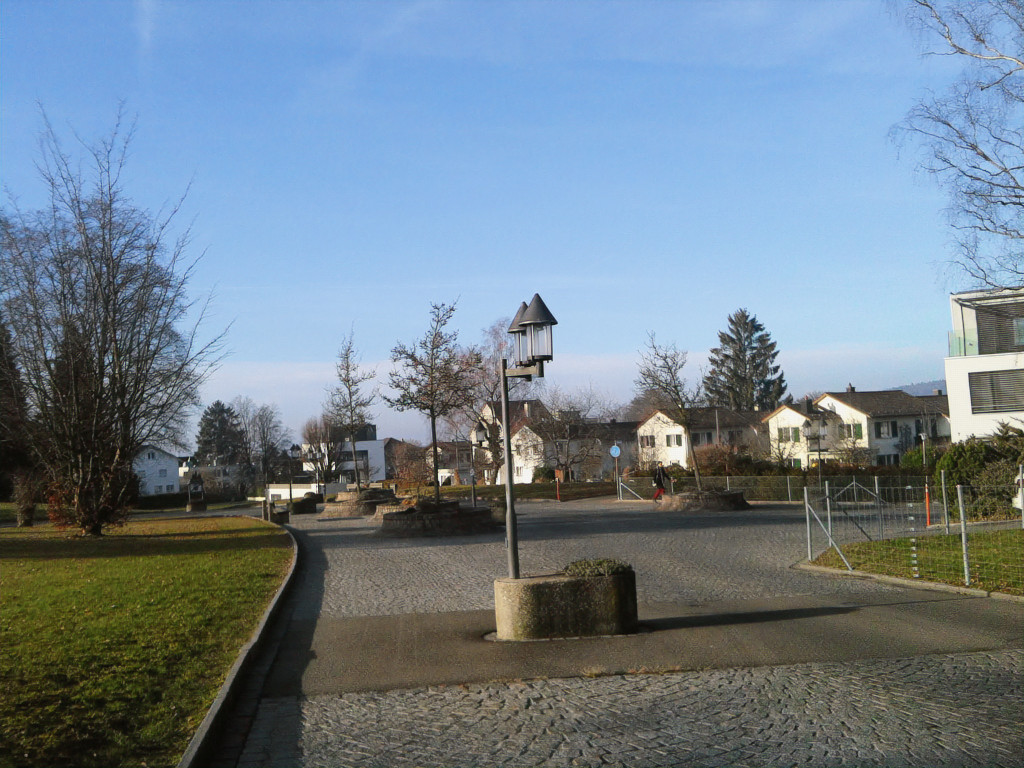}\\
   \includegraphics[width=0.497\linewidth]{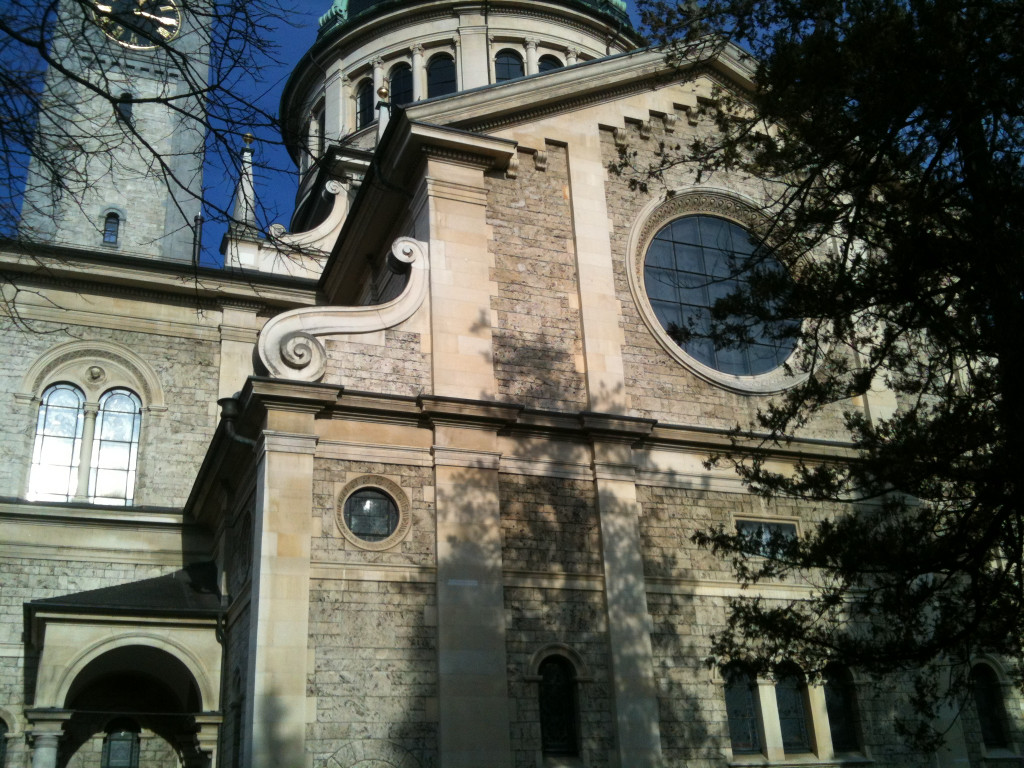}&
     \includegraphics[width=0.497\linewidth]{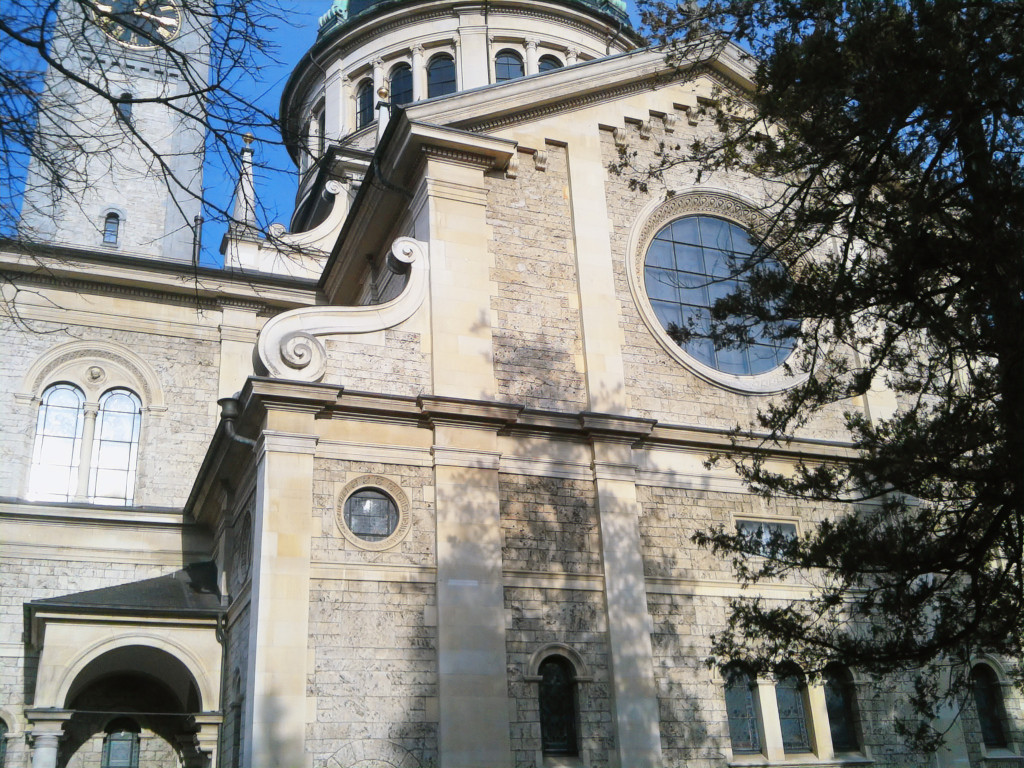}\\
   \includegraphics[width=0.497\linewidth]{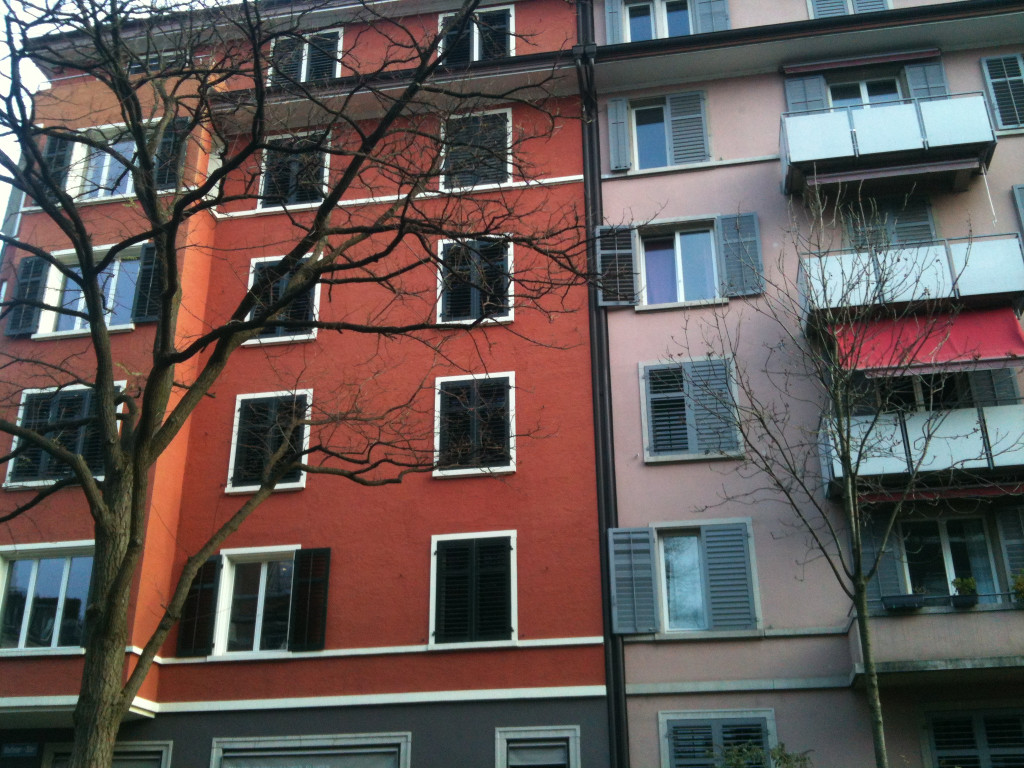}&
     \includegraphics[width=0.497\linewidth]{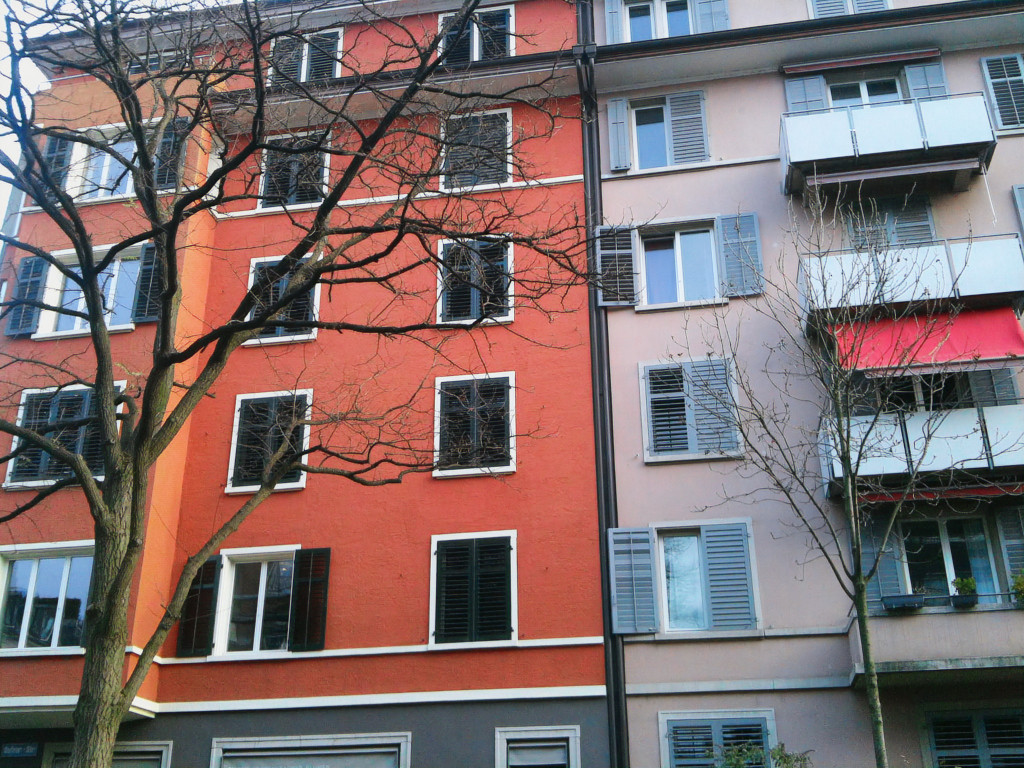}\\
\end{tabular}
\caption{Visual results for our method.}
\end{figure}

\end{document}